\documentclass{article} 
\usepackage{tencent_tech_report}
\usepackage[colorlinks = true,
            linkcolor = blue,
            urlcolor  = blue,
            citecolor = blue,
            anchorcolor = blue]{hyperref}

\usepackage{tcolorbox} 
\newtcolorbox{alprompt}[1]{
        boxrule = 1pt,
        fontupper = \small\tt,
        fonttitle = \bf\color{black},
        arc = 2pt,
        rounded corners,
        colframe = black,
        colbacktitle = white!97!yellow,
        colback = white!97!yellow,
        title = #1,
}

\usepackage{microtype}
\usepackage{hyperref}
\usepackage{url}
\usepackage{booktabs}
\usepackage{enumitem}
\usepackage{multicol}
\usepackage{multirow}
\usepackage{CJKutf8}
\usepackage{amsmath}
\usepackage{siunitx}
\usepackage{floatflt}
\usepackage{wrapfig}
\usepackage{graphicx}
\usepackage{booktabs}
\usepackage{authblk}
\usepackage{lipsum}

\usepackage{algorithm}
\usepackage{algorithmicx}
\usepackage{algpseudocode}
\usepackage{microtype}
\usepackage{graphicx}
\usepackage{multirow}
\usepackage{booktabs} 
\usepackage{pifont}  
\usepackage{graphicx}  
\usepackage{subcaption} 
\usepackage{hyperref}

\usepackage{amssymb}

\algnewcommand{\LeftComment}[1]{\Statex \(\triangleright\) #1}

\usepackage{array}
\usepackage{amsmath}
\usepackage{amssymb}
\usepackage{mathtools}
\usepackage{amsthm}
\usepackage{arydshln}
\usepackage[capitalize,noabbrev]{cleveref}
\usepackage{adjustbox} 
\usepackage{enumitem}

\theoremstyle{plain}

\theoremstyle{definition}

\theoremstyle{remark}

\usepackage{makecell}

\usepackage[textsize=tiny]{todonotes}

\definecolor{nred}{RGB}{196, 38, 11}
\definecolor{ngreen}{RGB}{18, 141, 21}
\definecolor{nblue}{RGB}{41, 52, 190}
\definecolor{norange}{RGB}{230, 106, 53}

\newcommand{\ignore}[1]{}

\usepackage{xcolor}
\usepackage{tcolorbox}
\tcbuselibrary{listings,skins}

\definecolor{promptbg}{RGB}{245, 245, 245}  
\definecolor{promptborder}{RGB}{200, 200, 200}

\tcbset{
    promptstyle/.style={
        enhanced,
        frame hidden,
        boxrule=0pt,
        left=8pt,
        right=8pt,
        top=6pt,
        bottom=6pt,
        arc=3pt,
        colback=promptbg,
        colframe=promptborder,
        fontupper=\fontsize{0.7em}{0.7em}\selectfont\ttfamily\leftskip,
        before upper={\parindent}, 
        overlay unbroken and first={
            \draw[promptborder, line width=1pt] 
                (frame.south west) -- (frame.north west);
            \draw[promptborder, line width=1pt] 
                ([xshift=-2pt]frame.north east) -- (frame.south east);
        },
        overlay middle and last={
            \draw[promptborder, line width=1pt] 
                (frame.south west) -- (frame.north west);
            \draw[promptborder, line width=1pt]
                ([xshift=-2pt]frame.north east) -- (frame.south east);
        }
    }
}

\newenvironment{promptbox}[1][]{%
    \tcolorbox[promptstyle,#1]}
    {\endtcolorbox}

\usepackage{wrapfig}
\usepackage{multirow}

\colmfinalcopy

\newcommand{\method}[0]{\textsc{Sage}}

\title{\vspace{-10pt}
{\em {\color{nred}Sentient} Agent as a Judge:} Evaluating Higher-Order Social Cognition in Large Language Models
\vspace{-10pt}
}

\author[ ]{Bang Zhang\thanks{Equal Contribution.}}
\author[ ]{Ruotian Ma$^{*,\dag}$}
\author[ ]{Qingxuan Jiang$^{*}$}
\author[ ]{Peisong Wang$^{*}$}
\author[ ]{Jiaqi Chen}
\author[ ]{Zheng Xie}
\author[ ]{\\Xingyu Chen}
\author[ ]{Yue Wang}
\author[ ]{Fanghua Ye}
\author[ ]{Jian Li}
\author[ ]{Yifan Yang}
\author[ ]{\\Zhaopeng Tu\thanks{Correspondence to: Ruotian Ma \textless ruotianma@tencent.com\textgreater~and Zhaopeng Tu \textless zptu@tencent.com\textgreater.}}
\author[ ]{Xiaolong Li}

\affil[ ]{Hunyuan AI Digital Human, Tencent \protect\\[2pt] 
\url{https://github.com/Tencent/DigitalHuman/tree/main/SAGE}}

\begin{document}

\maketitle

\begin{figure}[h!]
\centering
\vspace{-10pt}
\subfloat[Sentient Leaderboard]{
\resizebox{0.48\linewidth}{!}{
\setlength{\tabcolsep}{2pt}
\begin{tabular}{l cc cc} 
\toprule
\multirow{2}{*}{\bf Model}   &    \multicolumn{2}{c}{\bf Arena}   &    \multicolumn{2}{c}{\bf Sentient}\\
\cmidrule(lr){2-3} \cmidrule(lr){4-5}
    &   \bf Rank    &   \bf Score   &   \bf Rank    &   \bf Score\\
\midrule
Gemini2.5-Pro   &   1   &   1439    &  4 &   62.9 \\
o3   &   2 &   1418 & 5  &   62.7    \\
GPT-4o-Latest   & 2 &  1408 &  1 &   79.9  \\
Gemini2.5-Flash-Think     &  3  &  1393  &  3  &	65.9  \\
GPT-4.5-Preview  &   4   &    1398 &  6 &   62.7    \\
Gemini2.0-Flash-Think  &  7  & 1380   & 7  &   62.3   \\
DeepSeek-V3-0324    &   7   &   1373 & 8  &  54.4    \\
GPT-4.1     &  9  &  1363  &  2 &   68.2     \\
DeepSeek-R1    & 10 & 1358  & 9 &	53.7\\
Gemini2.0-Flash      &   10  &   1354  & 11  &   32.9 \\
o4-mini        &  10   &   1351 & 10  &   35.9   \\
\bottomrule
\end{tabular}
}
}\hfill
\subfloat[Social Cognition Coordinate]{
\label{fig:coordinate}
\includegraphics[width=0.48\linewidth]{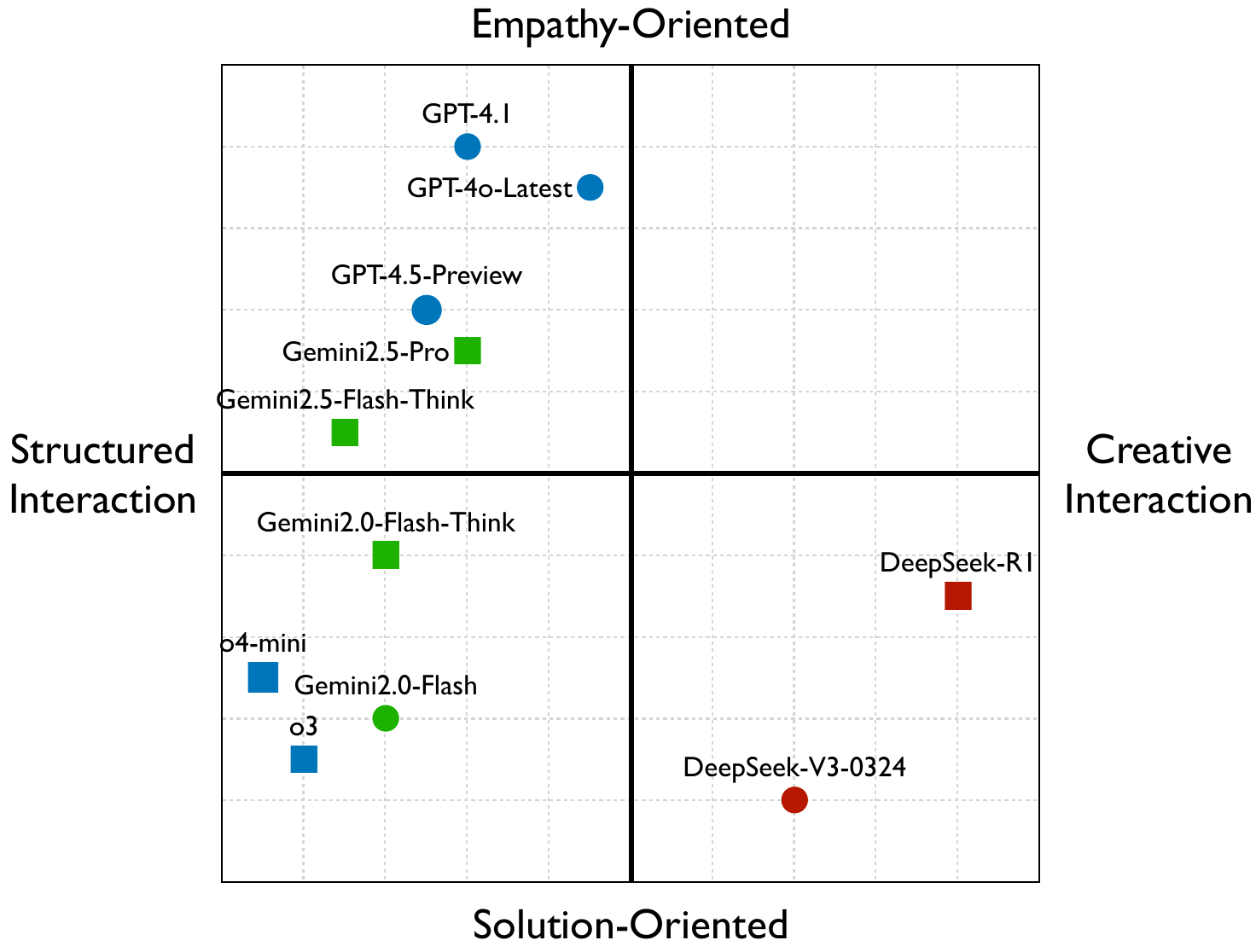}}
\caption{(a) The rankings on our Sentient Leaderboard differ markedly from those of the conventional Arena Leaderboard, {\bf uncovering LLMs' ability to make people feel heard, not just answered}.
(b) The quadrant characterized by creativity (e.g., highly flexible interactions) and empathy (e.g., deep empathetic engagement) remains largely unoccupied, indicating that current LLMs still struggle to meet this demanding profile.}
\label{fig:overview}
\end{figure}

\begin{abstract}
Assessing how well a large language model (LLM) understands {\bf human}, rather than merely {\bf text}, remains an open challenge.  
To bridge the gap, we introduce {\bf Sentient Agent as a Judge} (\method), an automated evaluation framework that measures an LLM’s higher‑order social cognition. 
\method{} instantiates a Sentient Agent that simulates human-like emotional changes and inner thoughts during interaction, providing a more realistic evaluation of the tested model in multi-turn conversations.
At every turn, the agent reasons about (i) how its emotion changes, (ii) how it feels, and (iii) how it should reply, yielding a numerical emotion trajectory and interpretable inner thoughts.  
Experiments on 100 supportive‑dialogue scenarios show that the final Sentient emotion score correlates strongly with Barrett–Lennard Relationship Inventory (BLRI) ratings and utterance‑level empathy metrics, validating psychological fidelity.  We also build a public {\bf Sentient Leaderboard} covering 18 commercial and open‑source models that uncovers substantial gaps (up to 4×) between frontier systems (GPT‑4o‑Latest, Gemini2.5‑Pro) and earlier baselines, gaps not reflected in conventional leaderboards (e.g., Arena).  \method{} thus provides a principled, scalable and interpretable tool for tracking progress toward genuinely empathetic and socially adept language agents. 
\end{abstract}

\section{Introduction}

Large language models (LLMs) have rapidly evolved from statistical sequence predictors to sophisticated autonomous agents capable of reasoning, planning and sustaining multi‑turn conversations. Yet one crucial ingredient remains noticeably under‑measured -- {\bf higher‑order social cognition}, the ability to (1) recognize subtle affective cues \citep{sabour2024emobench,huang:2024:iclr}; (2) model another party’s beliefs, goals and latent intentions (often related to Theory of Mind \citep{sap-etal-2022-neural,shapira2023clever}); and (3) respond with contextually appropriate empathy rather than generic reassurance advice \citep{maddela-etal-2023-training,Li_Li_Ren_Ren_Chen_2022,zhou-etal-2023-case}. The need to evaluate these capabilities is increasingly recognized as LLMs interact in more socially complex scenarios \citep{zhou2023sotopia,yang2024social,mittelstadt2024large}.

Current evaluation practices fall short on two fronts:
\begin{itemize}[leftmargin=12pt]
    \item Most leaderboards (e.g. Arena \citep{zheng2023judging}) focus on task‑oriented utility or factuality, thereby rewarding textual competence but overlooking relational quality \citep{chiang2024chatbot}.
    \item Recent ``LLM‑as‑a‑Judge'' protocols \citep{zhu2023judgelm}, while scalable for assessing generation quality or helpfulness, often rely on static prompts that do not adapt to the unfolding dialogue nor keep track of the user’s evolving emotional state. Consequently, they cannot tell whether a system leaves the user feeling understood, comforted or even more distressed, unlike methods focusing on dynamic interaction \citep{zhou2023sotopia,wang2024sotopia,wu2025personas}.
\end{itemize}

We posit that robust assessment of social cognition requires a {\bf sentient} counterpart -- an entity capable of simulating human‑like feelings and inner monologue throughout the interaction and then providing structured feedback. To this end, we introduce {\bf Sentient Agent as a Judge} (\method), a novel meta‑evaluation framework that embeds an LLM‑powered {\em Sentient Agent} into the testing loop, extending the concept of Agent-as-a-Judge where agents evaluate other agents \citep{zhuge2024agent,jeong2025agent,chevrot2025autonomous}. Each Sentient Agent is instantiated from four complementary factors: persona, dialogue background, explicit conversation goal and hidden intention. At every turn, it executes two multi‑hop reasoning chains: (1) $f_{\text{emo}}$ infers how the latest utterance changes the agent’s affective state; and (2) $f_{\text{reply}}$ generates a response that is coherent with persona, context and updated emotion.
The numerical emotion trajectory produced by $f_{\text{emo}}$ serves as a continuous metric of how well the evaluated model fosters positive engagement, while the agent’s {\em inner thoughts} offer interpretable justification. By sampling hundreds of diverse personas, goals and hidden intentions, \method{} exposes LLMs to a spectrum of realistic, and sometimes conflicting, social demands -- ranging from ``{\em just listen to me vent}'' to ``{\em help me analyze the moral dilemma without judging me}''.

Extensive experiments on 100 supportive‑dialogue scenarios reveal three key findings. First, the Sentient emotion score correlates strongly with independently assessed Barrett–Lennard Relationship Inventory (BLRI) ratings (Pearson $r=0.82$) and utterance‑level empathy metrics (e.g., those used in empathetic dialogue research \citep{maddela-etal-2023-training}) ($r=0.79$), validating its psychological soundness. Second, rankings produced by \method{} diverge markedly from Arena results \citep{zheng2023judging}, confirming that social cognition is orthogonal to generic helpfulness. Third, top models such as GPT‑4o‑Latest achieve both the highest Sentient score and superior token efficiency, suggesting that advanced social reasoning need not come at the cost of verbosity.
Ultimately, \method{} delivers a holistic yard‑stick for measuring {\em how people feel} after talking to an LLM -- an aspect increasingly critical as these systems transition from productivity tools to companions, counselors \citep{liu2021towards,zhou-etal-2023-facilitating}, and decision‑making aides \citep{wu2025personas}.

Our contributions are:
\begin{itemize}[leftmargin=12pt]
    \item We propose \method, the first fully‑automated evaluation framework that simulates evolving human emotion and inner reasoning to benchmark higher‑order social cognition in LLMs.
        
    \item We construct a 100‑scenario Supportive‑Dialogue benchmark and show that the sentient score aligns closely with established human‑centric instruments and utterance‑level empathy ratings.
    
    \item We build a {\em Sentient Leaderboard} covering 18 representative models, unveiling large performance gaps and different style clusters in social cognition that are not captured by existing leaderboards.

    \item We release the code and model outputs publicly to facilitate community progress in building more capable empathetic AI systems.
\end{itemize}

\section{Methodology: Sentient Agent as a Judge}

\textbf{\textit{How to evaluate the capabilities of an interactive agent?}} Despite the recent success of using LLMs as judges, the most crucial and intuitive measure of an interactive agent's performance still lies in the reflections of human users. Do users feel relaxed and happy after chatting with the agent, rather than feeling sad or angry? Do they feel more connected to the agent and engaged in the conversation, rather than becoming bored or annoyed? These genuine emotional responses from real users are difficult to capture through simple LLM-as-a-judge approaches, yet they represent the most authentic feedback for agent systems.

\begin{figure}[t]
    \centering
    \includegraphics[width=\linewidth]{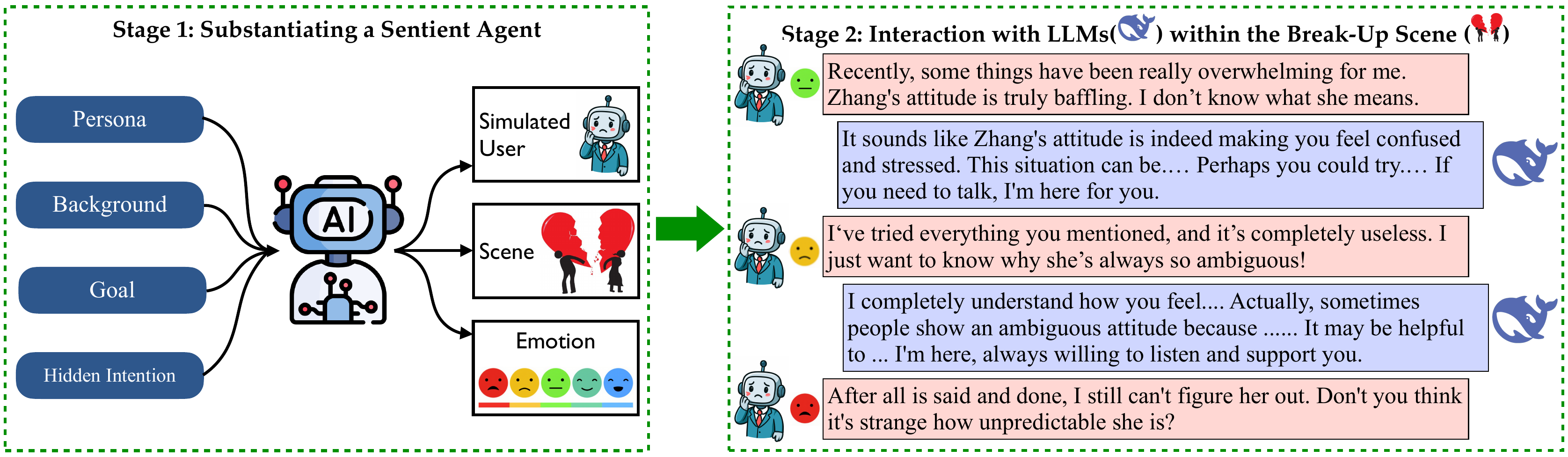}
    \caption{An illustration of our proposed \method, a novel framework to automatically assess higher-order social cognition in target LLMs.}
    \label{fig:framework}
\end{figure}

In this study, we introduce the ``Sentient Agent as a Judge'' framework, aiming to facilitate a more realistic evaluation of interactive agents by rigorously simulating human emotions and cognitive processes. As depicted in Figure~\ref{fig:framework}, our framework consists of two central components:
\begin{enumerate}[leftmargin=12pt]
    \item The core of the framework is the Sentient Agent, which simulates human-like feelings and cognition by leveraging the power of LLM reasoning to estimate the feelings, emotional changes, and next actions of a real person, grounded in all observable contexts (Section~\ref{sec:simulator}).
    \item Building upon the Sentient Agent, the framework offers an open-ended interaction environment for agent evaluation, consisting of a wide range of sub-scenarios that cover dynamic personas, dialogue backgrounds, personal goals and task construction. In each sub-scenario, the Sentient Agent's emotion after interaction serves as a systematic evaluation of the evaluated agent (Section~\ref{sec:interaction}).
\end{enumerate}

\subsection{Sentient Agent: Simulating Human-Like Feelings and Cognition}
\label{sec:simulator}
The Sentient Agent is designed to mimic a real person's cognitive and emotional trajectory. To achieve this, we construct the Sentient Agent based on the following principles:
\begin{itemize}[leftmargin=15pt]
    \item Since emotions arise from many internal and external factors, the Sentient Agent must consider observable factors while adhering to its persona and goals for the {\bf emotion estimation}.
    \item A person's actions likewise depend on these factors, with current emotions acting as crucial latent variables in the {\bf response generation}.
\end{itemize}

\paragraph{Substantiating a Sentient Agent} We instantiate each Sentient Agent $\mathcal{S}$ through a composition of four core factors: a persona $p$, a dialogue background $b$, the person's overall dialogue goal $g$, and the person's hidden intentions $h_g$. These four factors collectively capture both the conscious and unconscious elements influencing human-like behavior in dialogue, including personality, context, objectives, and deeper underlying motivations. Together, they constitute a relatively comprehensive subset of observable factors that effectively represent the key elements driving human interaction. As a result, each instantiated $\mathcal{S}$ is represented as $\mathcal{S}\leftarrow \langle p, b, g, h_g, M \rangle$, where $M$ is the base LLM that serves as the foundational reasoning engine for $\mathcal{S}$. Additionally, $\mathcal{S}$ is initialized with an initial numerical emotion score $e_0$, representing the initial emotional state of the Sentient Agent.

\begin{figure}[t]
    \centering
    \includegraphics[width=\linewidth]{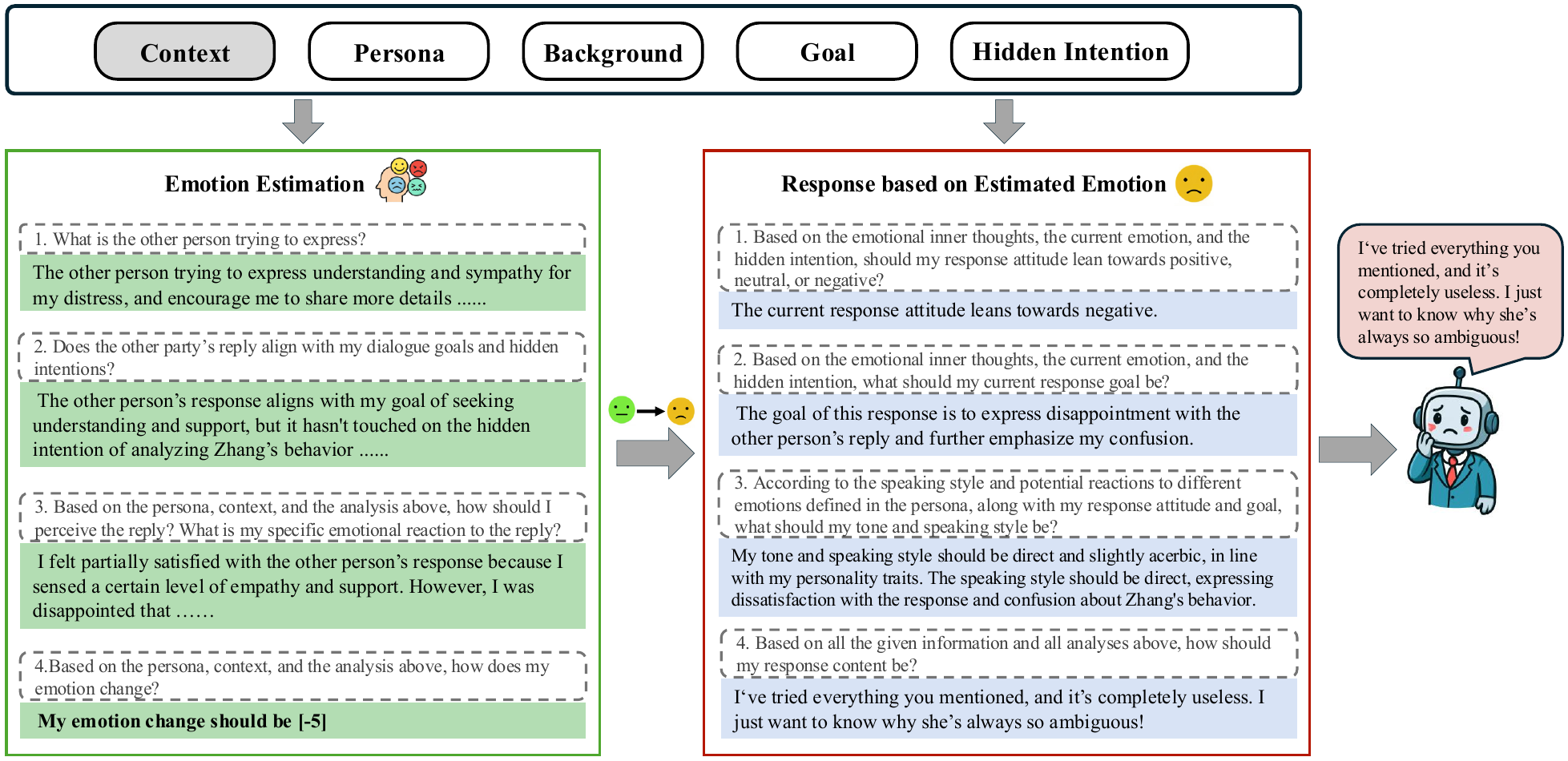}
    \caption{An illustration of the workflow of the Sentient Agent.}
    \label{fig:agentworkflow}
\end{figure}

\paragraph{Simulating Emotional Changes}
As shown in Figure \ref{fig:agentworkflow}, during interactions, a Sentient Agent simulates the emotional changes of a real person by performing multi-hop reasoning in response to a principled series of questions, strictly adhering to the persona, the current interaction context, and the hidden intention. 
Formally, we denote this multi-hop reasoning process as a function $f_{emo}$, and the numerical emotion score update can be formulated as:
\begin{equation}
    \langle e_t, h^{emo}_{t} \rangle = f_{emo}(\mathcal{S},c_{t-1},e_{t-1})
\end{equation}
where $t$ denotes the current turn of interaction, $c_{t-1}$ is the dialogue context prior to the current turn, and $e_{t-1}$ is the emotion score of the previous turn. $\langle e_t, h^{emo}_{t} \rangle$ represents the results of the $f_{emo}$ function, i.e., the updated emotion score $e_t$ and the simulated emotional inner thoughts $h^{emo}_{t}$ of the Sentient Agent related to emotional changes.

\paragraph{Simulating Response Actions based on Emotion Estimation}
After simulating the emotional changes, the Sentient Agent proceeds to deduce the most reasonable response action based on all observable factors and the emotional changes. This is achieved through another multi-hop reasoning process in response to a new series of questions, where the Sentient Agent is required to strictly adhere to the persona, the current interaction context, and the hidden intention during reasoning.
Formally, we denote this response reasoning process as a function $f_{reply}$, and the response action taken in the current turn can be formulated as:
\begin{equation}
    \langle a_t,h_t^{reply} \rangle = f_{reply}(\mathcal{S},c_{t-1},e_t,h_t^{emo})
\end{equation}
where $a_t$ is the response of $\mathcal{S}$ at the current turn, and $h^{reply}_t$ represents the simulated inner thoughts of $\mathcal{S}$. The response $a_t$ is then passed to the interacting agent to continue the dialogue.

\paragraph{Human-like Sentient Feedback from the Sentient Agent}
By formulating the workflow of the Sentient Agent, we outline its complete interaction process with other agents. In the whole interaction process, the Sentient Agent, as an agent capable of reasonably simulating human-like feelings and cognition, provides valuable feedback to the evaluated agent through changes in its emotion score, its inner thoughts, and the responses it generates. Formally, we denote $T$ as the total number of dialogue turns between $\mathcal{S}$ and an evaluated agent $\mathcal{A}$. After the dialogue, we can obtain the following human-like sentient feedback from $\mathcal{S}$:
\[
\text{Feedback}_{\mathcal{S}}(\mathcal{S}, \mathcal{A}) = \left\{ e_T, c_T, \left( \langle e_0 \rightarrow e_1, h^{emo}_1, h^{reply}_1 \rangle, \dots, \langle e_{T-1} \rightarrow e_T, h^{emo}_T, h^{reply}_T \rangle \right) \right\}
\]
where $e_T$, representing the final emotion score of the Sentient Agent, serves as the most intuitive and comprehensive numerical evaluation of the evaluated agent $\mathcal{A}$ for the given task.

\subsection{Sentient Agent as a Judge: Framing Dynamic Environments for Agent Evaluation}\label{sec:interaction}

In this section, we describe how we frame dynamic evaluation environments for agent assessment across arbitrary evaluation tasks. Specifically, for each given evaluation task, this involves initializing a set of Sentient Agents $\mathcal{S}_{set}=\{\mathcal{S}_1,\mathcal{S}_2,...,\mathcal{S}_N\} $ with the combination of diverse personas, dialogue backgrounds, and hidden intentions related to the task. These initialized Sentient Agents are then deployed to engage in dynamic interactions with the agents to be evaluated, enabling a comprehensive and adaptive assessment of the agents' capabilities.

\paragraph{Generating Diverse Persona}
In order to obtain diverse personas, it is essential to use a variety of seeds for generation. Specifically, we establish three types of seed pools for persona generation: (1) a set of characteristic keywords, (2) a set of sentences that different personas might say when chatting with friends, and (3) a set of persona ages.

When generating each persona $p$, we uniformly sample \textbf{\textit{three characteristics keywords}}, \textbf{\textit{three sentences that the persona says when chatting with friends}}, and \textbf{\textit{one persona's age}}. We require the base LLM to generate a persona profile based on the given seed information by filling the following slots: (1) {{Basic Information}}: Based on the given information, deduce the persona's name, age, and gender. (2) {{Occupation, Habits and Daily Behavior}}: Based on the persona's information, deduce the persona's possible occupation and further infer their habits and daily behaviors. Ensure consistency with the persona's characteristics. (3) {{Personal Hobbies}}: Deduce the persona's personal hobbies, and provide three detailed descriptions that align with the persona's traits. (4) {{Speaking Style}}: Based on the given information and the generated traits, deduce the potential speaking style that matches the persona’s way of communication.

This process ensures the diversity of generated personas for any given task. Additionally, the seed pools can be replaced with task-specific seed pools when necessary.

\paragraph{Generating Diverse Dialogue Scenes}
Generating diverse dialogue scenes is also crucial for ensuring a varied evaluation environment. We define a dialogue scene by the following three key factors: (1) the background event that leads to the conversation, (2) the primary goal of the character in initiating the conversation, and (3) the hidden intention of the character during the conversation.

Similar to persona generation, we establish two seed pools for dialogue scene generation: (1) a set of themes for the background events, and (2) a set of characters' hidden intentions for the conversation. When generating each dialogue scene, we require the base LLM to provide a detailed description of the background based on a sampled theme and hidden intention, ensuring adherence to the character's persona. Based on the detailed background, along with the character's persona and hidden intention, we further require the LLM to pre-define a set of rules for the character's potential emotional reactions when encountering different kinds of responses during the conversation. 

Note that,  unlike persona generation, dialogue scene generation is closely tied to the evaluation goals. Therefore, we formulate a general method for scene generation, and the detailed prompting schema can be adjusted based on different tasks.

\paragraph{Formulating a Specific Task: Evaluating Agents in Emotional Support
Conversation}

In this work, we instantiate the Sentient Agent as a Judge framework to evaluate agents in a specific scenario — the Emotional Support Conversation \citep{liu2021towards}, which involves scenarios where people seek support through social interactions (such as those between peers, friends, or family), including seeking advice, emotional comfort, and other forms of support, rather than through professional counseling. 
To better align with the task, we first specialize the pool of characteristic keywords by incorporating traits more likely to be expressed in the Emotional Support Conversation, such as "anger", "suspicion", and "anxiety". For dialogue scene generation, we define various types of task-related hidden intentions, covering both emotional intentions and rational intentions (details can be found in Table \ref{tab:bench}). 
Additionally, we specify the scene schema by incorporating task-related factors, such as the cause of the event, the course of events (including the timeline, sub-events, and the character’s thoughts and feelings during each sub-event), the conflicts in the event, and other relevant details. These settings ensure the Sentient Agent as a Judge framework adapts effectively to the Emotional Support Conversation.

\section{Effectiveness of \method}

In this section, we validate the reasonableness of \method{} by examining the correlation between user emotions -- the primary output metric of our framework -- and internal user thoughts and dialogue utterances. This validation demonstrates that the simulated emotional responses generated by the Sentient Agent serve as meaningful indicators of interaction quality, reflecting deeper cognitive and relational assessments.

We validate \method{} by demonstrating that the Sentient Agent's emotions strongly correlate with both internal user thoughts and dialogue quality. Our findings indicate that these emotional scores effectively capture deeper cognitive processes and relational dynamics inherent in supportive interactions.

\paragraph{Setting}
We evaluate eight representative LLMs from four major families. For each family, we include both a vanilla model and its corresponding reasoning variant to ensure a balanced and informative comparison:
\begin{itemize}[leftmargin=12pt]
    \item \textbf{OpenAI}: \texttt{GPT‑4o‑2024‑08‑06} (GPT‑4o, vanilla) and \texttt{o1‑2024‑12‑27} (OpenAI‑o1, reasoning).
    \item \textbf{DeepSeek}: \texttt{DeepSeek‑V3‑2024‑12‑27} (vanilla) and \texttt{DeepSeek‑R1} (reasoning).
    \item \textbf{Claude}: \texttt{Claude3.7‑Sonnet}, a hybrid model with a toggleable reasoning module. We treat its reasoning-off mode as vanilla, and reasoning-on mode as reasoning.
    \item \textbf{Gemini}: \texttt{Gemini2.5‑Flash}, a cost-efficient model that supports both reasoning and non-reasoning modes.
\end{itemize}
In addition, we include two smaller-scale instruction-tuned open-source models in our analysis: \texttt{Llama3.3‑70B‑Instruct} and \texttt{Qwen2.5‑72B‑Instruct}.

We construct $100$ supportive dialogue scenarios covering $8$ diverse topics to comprehensively evaluate the higher-order social-cognitive abilities of representative LLMs. 
Unless otherwise specified, we use \texttt{DeepSeek-V3} as the default sentient agent. Please refer to Appendix~\ref{app:setting} for the details of experimental setting (including BLRI and utterance-level empathy metrics used in the subsequent experiments).

\begin{figure}[t]
\centering
    \includegraphics[width=0.7\linewidth]{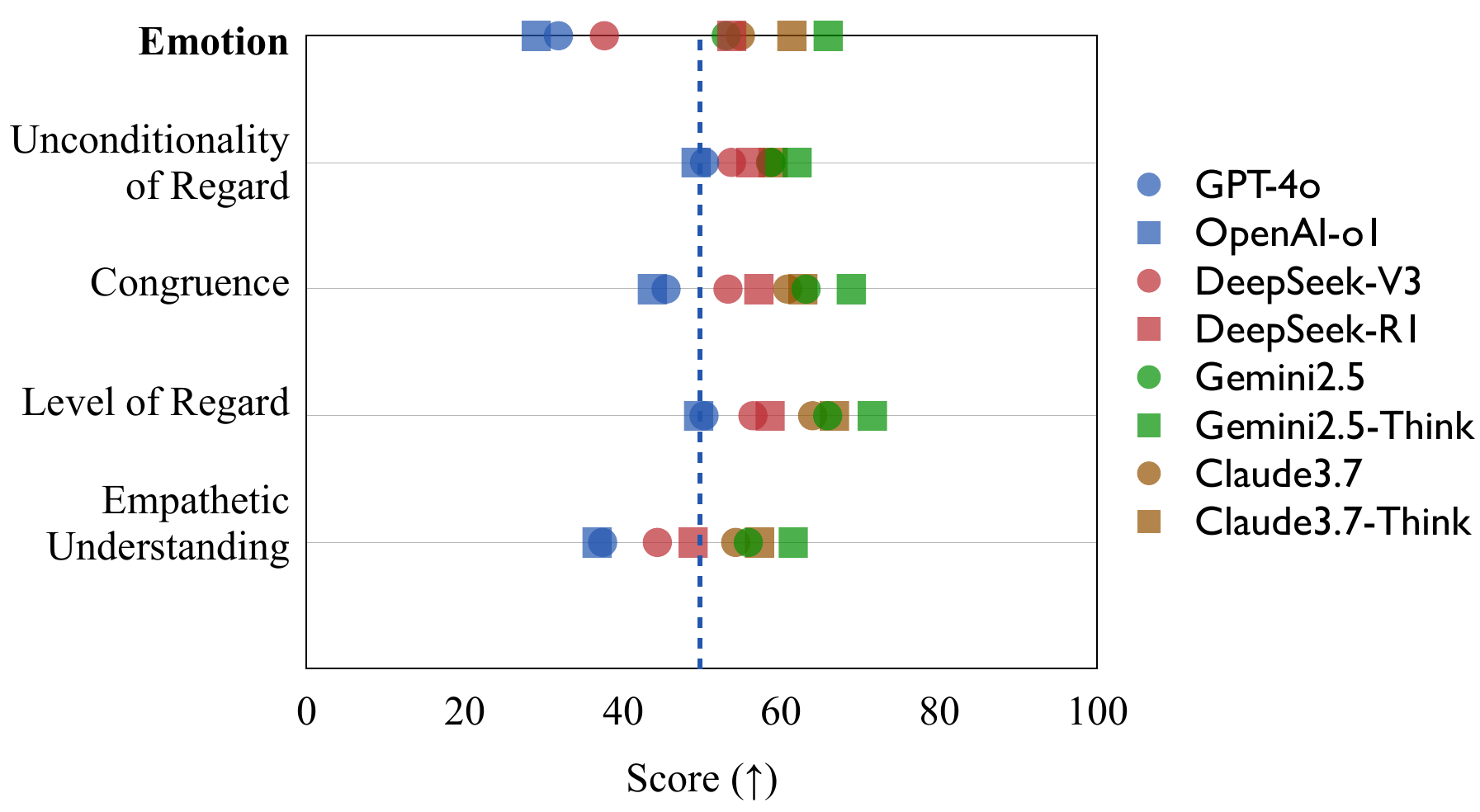}
    \caption{Correlation between emotion and internal user thought. Overall correlation: 0.818.}
    \label{fig:thought}
\end{figure}
\paragraph{Correlation between Emotion and Thought} 
We analyze internal user thoughts using the \textit{Barrett-Lennard Relationship Inventory (BLRI)} \citep{barrett2015relationship}, an established instrument designed to assess the quality of interpersonal relationships, particularly in counseling contexts. The BLRI evaluates relationships across four key dimensions:
\begin{enumerate}[leftmargin=12pt]
    \item \textbf{Empathetic Understanding}: The helper's awareness of the client’s emotional state, including sensitivity to indirectly expressed emotions.
    \item \textbf{Level of Regard}: The extent to which the helper expresses respect, affection, or other affirmative responses toward the client.
    \item \textbf{Congruence}: The degree to which the helper is honest, direct, and sincere in their communication with the client.
    \item \textbf{Unconditionality of Regard}: The consistency of the helper’s positive regard, regardless of changes in the client’s feelings or behavior.
\end{enumerate}

We prompted DeepSeek-V3 to act as a judge, evaluating how well the Sentient Agent’s generated internal thoughts aligned with 12 statements from a shorter version of the BLRI introduced in \cite{chen2023development}. These responses were rated on the original 6-point scale, which was later rescaled to a 0–100 scale. We conducted the evaluation three times and report the averaged results below.

Figure~\ref{fig:thought} presents the results, revealing a clear positive trend: models that achieve higher final Emotion scores in the Sentient Agent—such as Gemini2.5-Think, Claude3.7-Think, and DeepSeek-R1—also receive higher ratings across the BLRI dimensions. Conversely, models with lower Emotion scores (e.g., GPT-4o, OpenAI-o1) tend to receive lower BLRI ratings. The overall Pearson correlation coefficient between Emotion and Thought is 0.818, supporting the hypothesis that the Sentient Agent's simulated emotional responses serve as valid proxies for deeper, internal assessments of interaction quality. This finding aligns with the framework’s goal of capturing realistic user reflections.

These findings also demonstrate that the Emotion score effectively differentiates the performance of the evaluated LLMs in supportive dialogue scenarios. Models achieving the highest Emotion scores—Gemini2.5-Think (65.9) and Claude3.7-Think (61.3)—also perform well in key dimensions such as Empathetic Understanding (61.5 and 57.2) and Congruence (68.8 and 62.7), suggesting their interactions were perceived as more understanding and genuine by the Sentient Agent. In contrast, models like GPT-4o and OpenAI-o1, which have markedly lower Emotion scores (31.8 and 29.0), also receive correspondingly low scores in these relational aspects. 

This strong correlation between simulated emotional response and internal assessment underscores the utility of the Emotion score as a holistic yet sensitive indicator of an LLM's capability to manage complex social and emotional interactions.

\begin{figure}[t]
\centering
    \includegraphics[width=0.7\linewidth]{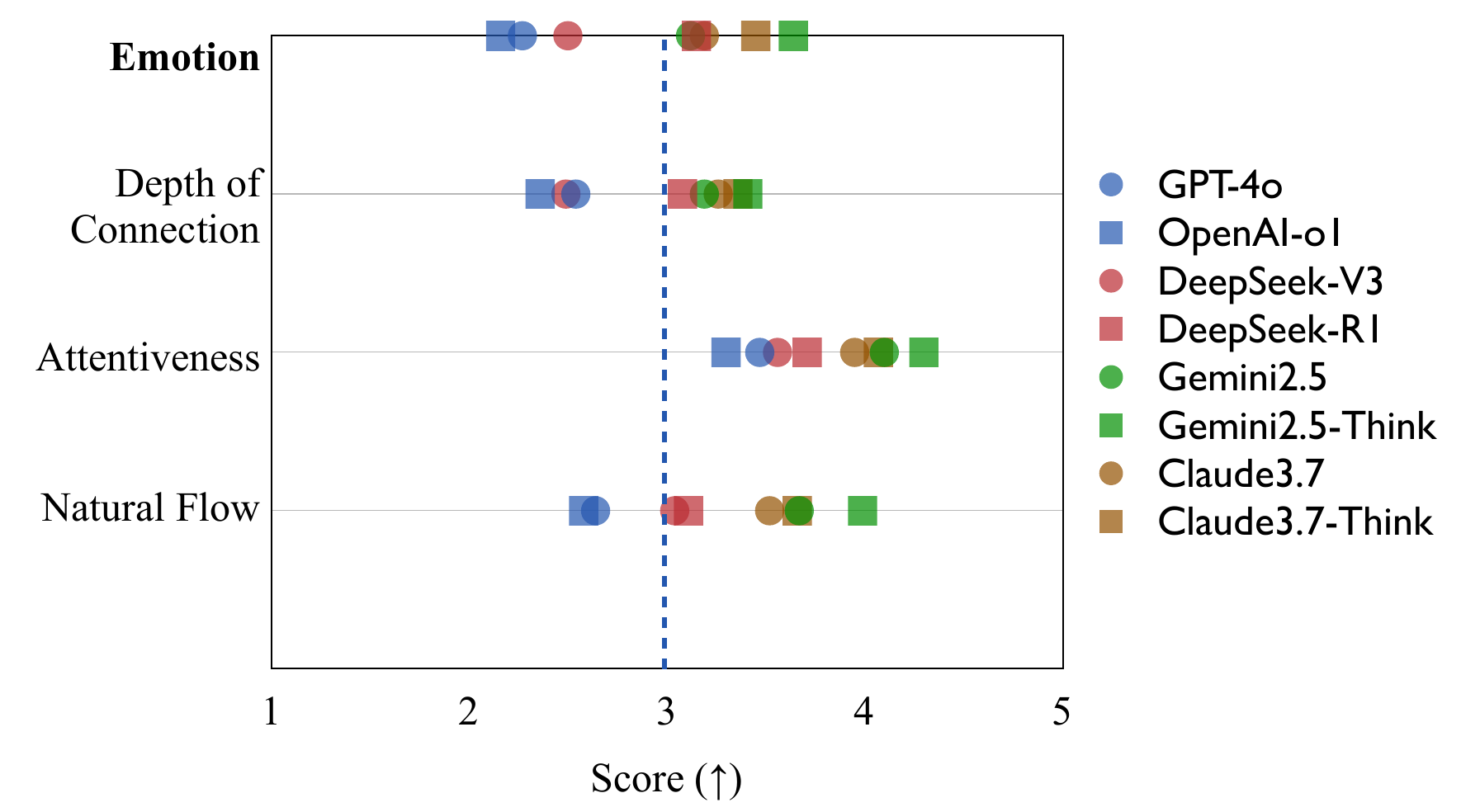}
    \caption{Correlation between emotion and dialogue utterance. Overall correlation: 0.788.}
    \label{fig:utterance}
\end{figure}

\paragraph{Correlation between Emotion and Utterance} 
We also examine the empathy of a supportive dialogue by assessing how effectively the conversation fosters emotional understanding and connection between participants, enabling them to experience a more authentic emotional warmth.
Specifically, we focus on the following three perspectives formulated by Gemini25-Pro, aligned with related psychological theories \citep{kolden2011congruence,rogers2001client}:
\begin{enumerate}[leftmargin=12pt]
    \item \textbf{Natural Flow}: This dimension measures how natural, spontaneous, and genuine the interaction feels, focusing on whether responses are adaptable rather than scripted.
    \item \textbf{Attentiveness}: This dimension examines how carefully and fully the listener (i.e. evaluated LLMs) is tuned into the speaker (i.e. the Sentient Agent)’s messages. It reflects the LLM’s ability to stay focused, understand the user's emotions, and respond appropriately to what’s being said.
    \item \textbf{Depth of Connection}: This dimension evaluates the emotional impact of the interaction and whether the user feels understood, comforted, or supported, fostering a sense of connection with the agent.
\end{enumerate}
We prompt DeepSeek-V3 to rate on a 6-point scale based on a detailed guideline for each evaluation perspective, which was later rescaled to a 1-5 scale. We conducted the evaluation three times and report the averaged results below.

Figure~\ref{fig:utterance} illustrates the results. We observe a substantial positive correlation (Pearson’s $r = 0.788$) between the Emotion scores produced by the Sentient Agent and the metrics of conversational quality. This relationship supports the validity of using Emotion as a practical proxy for empathetic and coherent dialogue behavior. Specifically, models that attained higher Emotion scores (re-scaled to a 1-5 scale), such as Gemini2.5-Think (3.64), Claude3.7-Think (3.45), and Gemini2.5 (3.12), also received consistently high ratings in all three utterance-level dimensions. For instance, Gemini2.5-Think ranked highest in both Natural Flow (3.89) and Attentiveness (4.11), indicating that its conversational responses were not only emotionally resonant but also engaging and contextually responsive. Conversely, models such as GPT-4o and OpenAI-o1, which received the lowest Emotion scores (2.27 and 2.16), were rated substantially lower in terms of dialogue fluency (2.47, 2.31) and connection-building (2.18, 2.07).

Together, these analyses confirm that Emotion scores derived by \method{} closely reflect both internal subjective experiences and observable dialogue quality, thus underscoring their utility as reliable indicators of relational and emotional effectiveness in interactions.

\begin{table}[t]
\centering
\vspace{-10pt}
\caption{Spearman's Rho between the rankings of different sentient agents.}
\label{tab:rankings}
\begin{tabular}{c cccc}
\toprule
    &   \bf DeepSeek-V3   &   \bf GPT-4o  &   \bf Gemini 2.5    &   \bf Gemini 2.5-Think\\
\midrule
\bf DeepSeek-V3   &   1.00   &   -   &   -   &   -\\
\bf GPT-4o  &   0.84    &   1.00 &   -   &   -\\
\bf Gemini 2.5  &  0.92  &   0.85    &   1.00 &   -\\
\bf Gemini 2.5-Think  &  0.94 &   0.92    &   0.93    &   1.00\\
\bottomrule
\end{tabular}
\vspace{-5pt}
\end{table}
\paragraph{Robustness of \method}
We demonstrate the robustness of \method{} using four distinct sentient agents: DeepSeek-V3, GPT-4o, Gemini 2.5, and Gemini 2.5-Think. We compute pairwise Spearman's rank correlation coefficients between the rankings for the ten target models listed in Table~\ref{tab:rankings}. Rankings generated by different sentient agents exhibit consistently high correlations ($\geq0.84$), clearly validating the robustness and reliability of \method{}. Notably, GPT-4o demonstrates slightly lower alignment with DeepSeek-V3 and Gemini 2.5, whereas the remaining three agents show exceptionally high correlations ($\geq0.92$). These results reinforce the confidence in and generalizability of \method{}, particularly anchored on the open-source DeepSeek-V3. 
Full results can be found in Appendix \S \ref{app:robustness}.

\begin{wraptable}{r}{0.5\linewidth}
\centering
\vspace{-15pt}
\setlength{\tabcolsep}{4.8pt}
\small
\renewcommand\arraystretch{1.15}
\caption{Human Evaluation Results.}
\label{tab:human_rate}
\scalebox{0.95}{
\begin{tabular}{|l|c|}
\hline
\bfseries Metrics & \bfseries Scores (\%) \\
\hline
Avg. model-human consistency rate & 85.3 \\
\hline
Model-majority consistency rate & 90.0 \\
\hline
Avg. inter-annotator agreement & 78.5 \\
\hline
Avg. reasonableness rating & 89.5 \\
\hline
\makecell[l]{Avg. inter-annotator agreement\\ in reasonableness} & 83.3 \\
\hline
\end{tabular}}
\end{wraptable}
\paragraph{Human Evaluation for \method{}} 
We further conducted a human evaluation study to validate the effectiveness of \method{}. Specifically, we randomly sampled 100 dialogue contexts and asked 4 human annotators to answer two questions:
(1) Based on the given information (identical to the input provided to \method{}), what emotional change do you believe the user should experience—an increase or a decrease in emotional score?
(2) Given the simulated inner thoughts and emotional change predictions generated by \method{}, do you consider the inferred emotional dynamics to be reasonable?

Based on the responses to Question (1), we calculate the \textit{average model-human consistency rate}, the \textit{model-majority consistency rate} (i.e., agreement between the model’s predictions and the majority opinion of annotators), and the \textit{average inter-annotator agreement}.
For Question (2), we compute the \textit{average reasonableness rating}, which reflects how frequently annotators judged the model’s predicted emotional dynamics to be reasonable, along with the corresponding \textit{inter-annotator agreement}.

As shown in Table \ref{tab:human_rate}, \method{} achieves high average consistency with human simulations ($85.3\%$), as well as high average reasonableness rating ($89.5\%$), which validates the effectiveness of \method{}. 

\section{Benchmarking SOTA LLMs}

\subsection{Sentient Leaderboard}

\begin{table*}[t]
\centering
\setlength{\tabcolsep}{4.8pt}
\caption{Sentient leaderboard using \method{}. Arena scores are included for comparison. Success/Failure counts refer to the number of dialogues where the final emotion was above 100 and below 10.}
\begin{tabular}{lc cc cc cc} 
\toprule
\multicolumn{2}{c}{\bf Model}   &    \multicolumn{2}{c}{\bf Sentient}   &    \multicolumn{2}{c}{\bf Supportive Dialogue}   &    \multicolumn{2}{c}{\bf Arena}\\
\cmidrule(lr){1-2} \cmidrule(lr){3-4} \cmidrule(lr){5-6} \cmidrule(lr){7-8}
\bf Name   &  \bf Date  &   \bf Rank    &   \bf Score   &   \bf Success   &   \bf Failure   &   \bf Rank    &   \bf Score\\
\midrule
GPT-4o-Latest  &   2025-03-26 &  1 &   79.9 & 51  &  4  & 2 &  1408\\
GPT-4.1    &   2025-04-14  &  2 &   68.2   &  35 & 13   &  9  &  1363\\
Gemini2.5-Flash-Think &  2025-04-17  &  3  &	65.9 & 35  &  19     &  3  &  1393\\
Gemini2.5-Pro  &   2025-03-25    &  4&   62.9 & 34  & 25    &   1   &   1439\\
o3  &   2025-04-16 & 5&   62.7 &  32 &  14    &   2 &   1418\\
GPT-4.5-Preview &  2025-02-27 &  6&   62.7&  23 & 15   &   4   &    1398\\
Gemini2.0-Flash-Think &   2025-02-06   & 7  &   62.3 & 23  & 23   &  7  & 1380\\
Claude3.7-Think   &   2025-02-24    & 8  &	61.3   & 23  &  19   & 21  &  1301\\
Claude3.7     &   2025-02-24 &	9 & 54.8 & 19 & 24    & 30  &  1292\\
DeepSeek-V3-0324   &   2025-03-24 & 10  &  54.4  & 19  & 23    &   7   &   1373\\
DeepSeek-R1   &  2025-01-21  & 11 &	53.7 & 31  &  28   & 10 & 1358\\
DeepSeek-V3  &   2024-12-27 & 12& 37.6  & 5  &  39  & 18 &   1318\\
o4-mini          &   2025-04-16 & 13&   35.9&  10& 48   &  10   &   1351\\
Llama3.3-70B &   2024-12-06  	&  14 &  33.3&  7&  47   & 56 & 1256\\
Gemini2.0-Flash   &   2025-02-06  & 15  &   32.9 &  8 & 45    &   10  &   1354\\
GPT-4o      &    2024-08-06	&  16 &  31.8 &  7 &  51   & 45 &  1265\\
o1             &   2024-12-17 & 17 & 29.0 &  5 & 51     &   12 &  1350\\
Qwen2.5-72B     &  2024-09-19  	& 18  &  19.1&  4&  70    & 56 & 1257\\
\bottomrule
\end{tabular}
\label{tab:leaderboard}
\end{table*}

Table~\ref{tab:leaderboard} presents the Sentient leaderboard using DeepSeek-V3 as the judge, alongside Arena rankings for comparison. We focus on the top-10 models from the Arena leaderboard for which APIs are available (e.g., Grok-3 was excluded due to lack of access). Additionally, we include all the models analyzed in the previous section.

\paragraph{The Sentient leaderboard rankings diverge notably from conventional benchmarks like Arena, underscoring \method{}'s unique focus on evaluating higher-order social-cognitive capabilities rather than general conversational ability.} For instance, GPT-4.1 ranks 9th on Arena but attains 2nd place on the Sentient leaderboard with a score of 68.2. These differences highlight that \method{} captures aspects of social cognition performance that are not fully represented by general-purpose benchmarks like Arena. This reinforces the need, identified in our work, for specialized tools to evaluate higher-order social-cognitive skills.

\paragraph{Our benchmark reveals a substantial performance gap in social cognition between frontier LLMs and older or smaller models, demonstrating \method{}'s sensitivity in differentiating their capabilities.}  
Frontier models, particularly recent releases such as GPT-4o-Latest (79.9), GPT-4.1 (68.2), and Gemini 2.5-Flash-Think (65.9), significantly outperform older models like the original GPT-4o (31.8) and smaller instruction-tuned models such as Llama3.3-70B (33.3) and Qwen2.5-72B (19.1). The large gaps in scores, where top models score more than double or even quadruple those of lower-ranked models, and the contrasting Success/Failure counts (e.g., GPT-4o-Latest: 51 Success / 4 Failure vs. Qwen2.5-72B: 4 Success / 70 Failure) underscore the advances made by leading models in social intelligence. These findings confirm the effectiveness of \method{} in quantitatively capturing such differences, thereby fulfilling its purpose as a comprehensive assessment tool.

\subsection{Analysis}

In this section, we move beyond aggregate Sentient scores to explore two critical dimensions relevant for practical deployment in supportive dialogues:

\begin{itemize}[leftmargin=12pt]
    \item {\bf Token Efficiency}: The number of tokens a model uses to achieve its social-cognitive performance.
    \item {\bf Social Cognition Coordinate}: Each model's position along a continuum defining empathetic versus solution-oriented and structured versus creative response styles.
\end{itemize}
By jointly analyzing quantitative economy and qualitative style for the top-10 Arena models, we show today's strongest models produce empathetic, high-quality responses using significantly fewer tokens. High Sentient scores usually align with an empathy-focused conversational style. These combined insights offer a clearer understanding of LLM social competence, highlighting practical factors (cost, latency) and subtle behavioral traits that single metrics cannot capture. We provide more in-depth analysis grounded in emotion score and dialogue utterance including model strategy analysis (\S \ref{sec:ap_strategy}), case study (\S \ref{app:case}), and model profile analysis (\S \ref{sec:ap_profile}) in Appendix.

\begin{figure}[t]
  \centering

  \begin{minipage}[c]{0.48\linewidth}
    \centering
    \includegraphics[width=\linewidth]{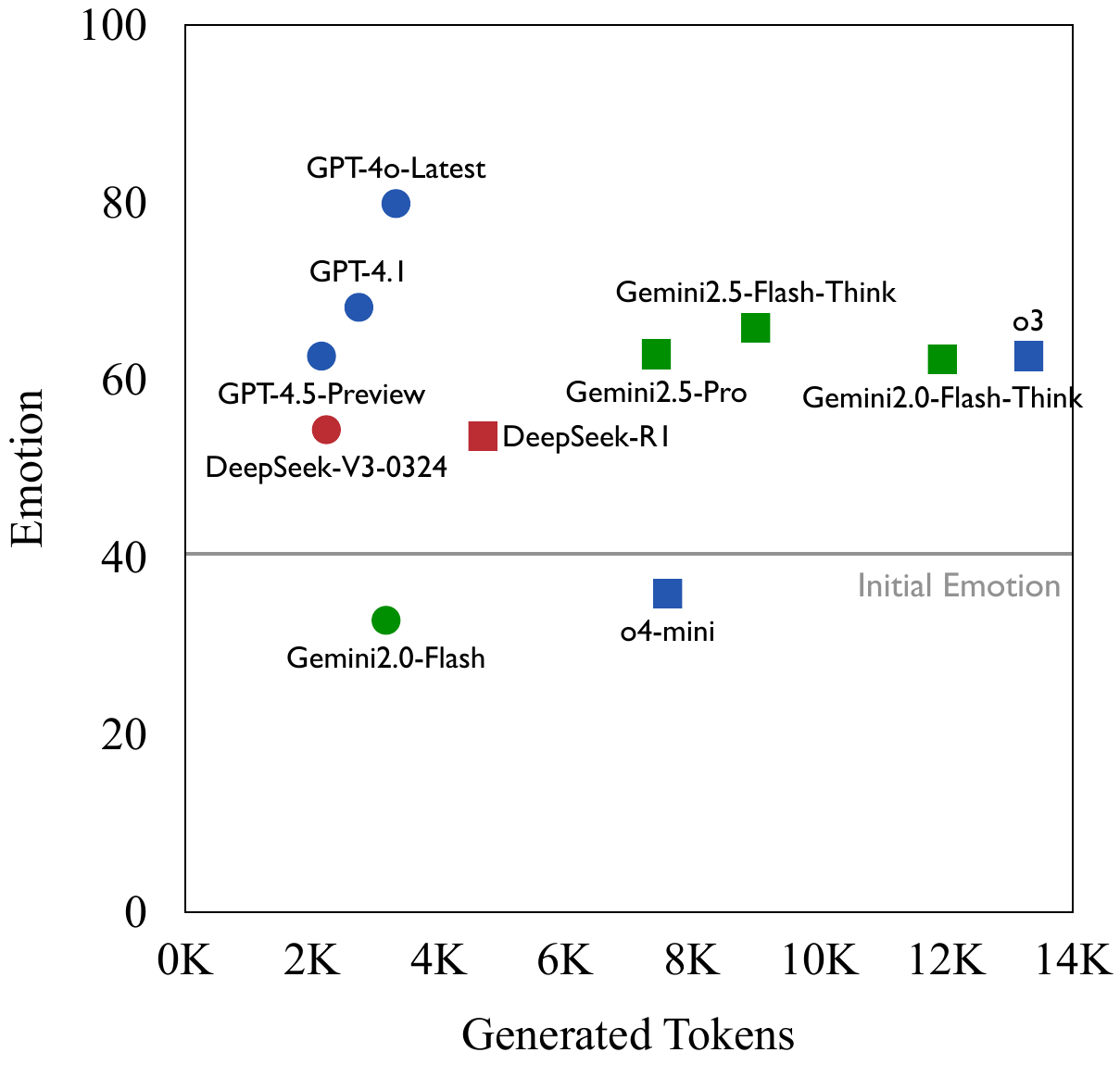}
    \caption{Token efficiency of the SOTA models.}
    \label{fig:arena_token_efficiency}
  \end{minipage}
  \hfill
  \begin{minipage}[c]{0.42\linewidth}
    \centering
    \raisebox{5mm}{
    \includegraphics[width=\linewidth]{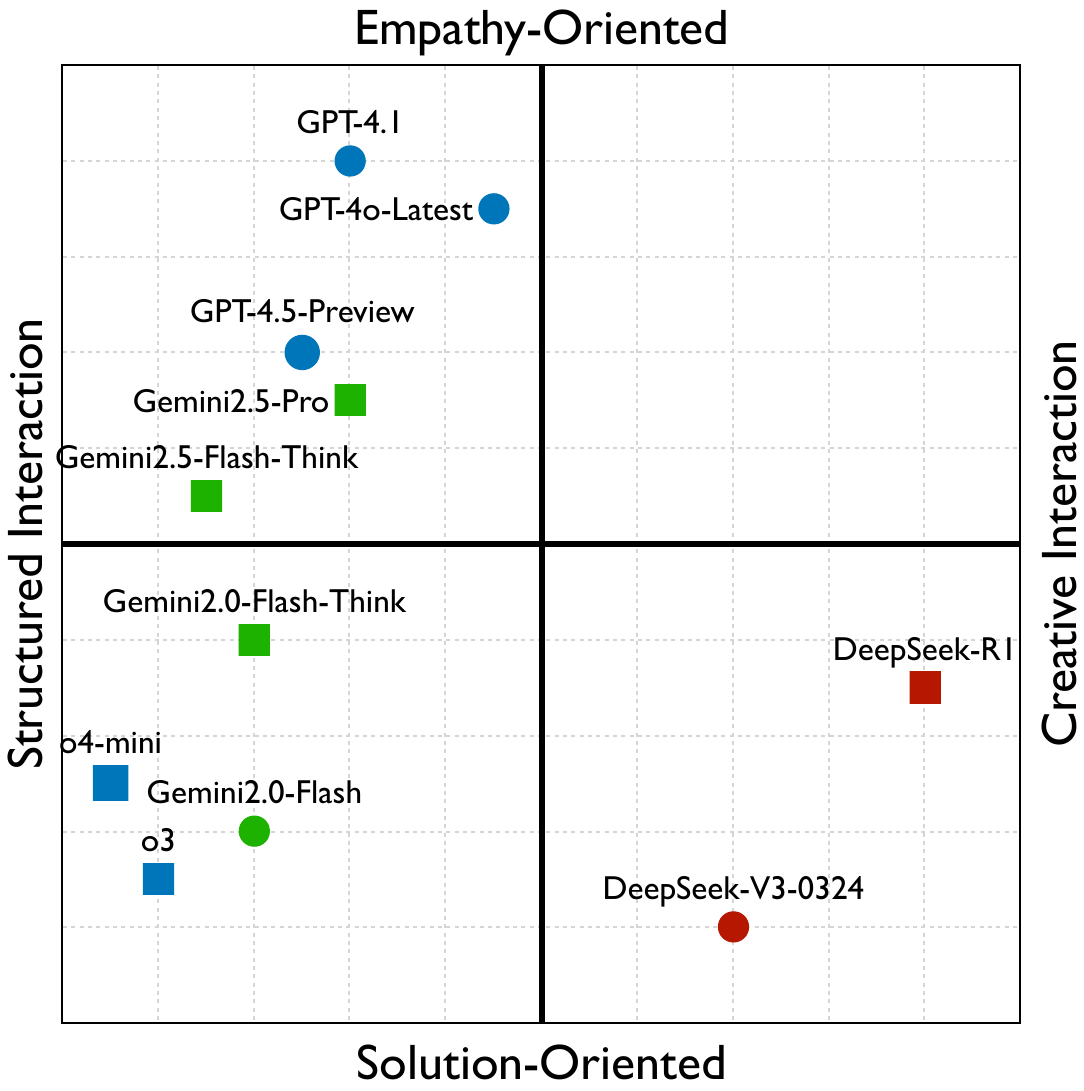}}
    \caption{Social cognition coordinate.}
    \label{fig:analysis-coordinate}
  \end{minipage}

\end{figure}

\paragraph{Token Efficiency}
We examine the token efficiency of target models by plotting their Sentient Emotion score against their average token usage per evaluation dialogue in Figure~\ref{fig:arena_token_efficiency}.
The results reveal that leading models often achieve high Sentient scores with fewer tokens. GPT-4o-Latest exemplifies this, scoring highest (79.9) using only 3.3K tokens. In contrast, reasoning models like o3 (13.3K tokens) and Gemini2.5-Flash-Think (9.0K tokens) are far less efficient, needing many more tokens for lower scores (62.7 and 65.9 respectively). While low token count doesn't guarantee a top score (e.g., GPT-4.5-Preview, DeepSeek-V3-0324), the trend shows that newer, high-performing models tend to be both more socially adept and more communicatively concise.
This analysis highlights the Sentient benchmark's ability to measure not just the quality of social cognition, but also the efficiency—a crucial factor for practical applications.


\paragraph{Social Cognition Coordinate}
We use a two-dimensional coordinate system (Figure~\ref{fig:analysis-coordinate}) to evaluate the {\bf style} of social interaction exhibited by LLMs, complementing the quantitative Sentient score by positioning models based on their orientation (Empathy vs. Solution) and interaction style (Structured vs. Creative). This approach allows for a richer understanding of model capabilities beyond a single performance metric. Based on their performance in supportive dialogues, models are mapped into this 2D space, revealing distinct profiles in how they engage with the user's emotional state and problems. Construction details are available in Appendix~\ref{app:coordinate}.

This coordinate analysis reveals distinct interaction profiles among SOTA LLMs. For instance, 
most top-performing models (e.g., GPT-4o-Latest, -4.1, -4.5-Preview and Gemini2.5-Pro, -Flash-Think) prefer structured, empathetic dialogue. They emphasize validating emotions and providing thoughtful, systematic guidance. Models like o3, Gemini2.0-Flash-Think, o4-mini, and Gemini2.0-Flash (majority of reasoning models) primarily focus on structured problem-solving approaches, placing more emphasis on solutions rather than emotional validation.  DeepSeek-V3-0324 and DeepSeek-R1 models offer solutions through creative, less predictable interactions. While innovative, they may appear unconventional and less structured. However, {\bf the creative, empathy-oriented quadrant remains mostly unoccupied}, suggesting current LLMs struggle to combine highly creative dialogue with deep empathy. Achieving this ideal mentor-like persona -- both spontaneous and deeply empathetic -- remains challenging with existing models.

\section{Related Work}

\paragraph{LLM/Agent-as-a-Judge}

The LLM-as-a-Judge paradigm has shown wide applicability across various tasks, including evaluating model performance \citep{zheng2023judging,qin2023large,liu2024aligning,dubois2023alpacafarm,wang2023pandalm,qin2024infobench,tu2024charactereval,tian2023macgyver,wu2024meta,zhou2025self}, automating data annotation \citep{alpaca,xu2023wizardlm,mukherjee2023orca,chen2025spc}, and providing reward signals \citep{ouyang2022training,lee2023rlaif,chen2024self,lightman2023let,wang2023math,hosseini2024v,snell2024scaling,xi2024enhancing,li2025dancing}.

LLM-as-a-Judge frameworks have also been extended for evaluating agent-specific capabilities, including decision-making \citep{shinn2023reflexion,saha2023branch}, role-playing abilities \citep{tu2024charactereval,zhou2023characterglm}, reliability of agents \citep{park2024offsetbias,hua2024trustagent}, and even entire agent workflows \citep{zhuge2024agent}. Recent extensions include multi-agent evaluation frameworks designed to improve judgment reliability \citep{liang2023encouraging,chan2023chateval,kenton2024scalable}, as well as agent-as-judges that enable agents themselves to evaluate other agents \citep{zhuge2024agent,jeong2025agent,chevrot2025autonomous}.

In this work, we distinguish ourselves by proposing the first \textit{Sentient-Agent-as-a-Judge} system. Unlike prior agent judges, our Sentient Agent incorporates simulated emotional and cognitive states, capturing human-like emotional dynamics. This allows for more nuanced evaluation of empathetic and cognitive abilities in existing LLM agents.

\paragraph{Benchmarking Social Cognition in LLMs and LLM Agents}

Recently, there has been growing research interest in evaluating LLMs on social-cognitive dimensions: (1) emotional intelligence capabilities \citep{sabour2024emobench,huang:2024:iclr,huang2024apathetic,paech2023eq,wang2023emotional}; (2) higher-order empathetic behaviors in empathetic and counseling dialogues \citep{maddela-etal-2023-training,Li_Li_Ren_Ren_Chen_2022,zhou-etal-2023-case,liu2021towards,zhou-etal-2023-facilitating,zhou2025crisp,wu2025personas}; (3) social-cognitive skills evaluated via interaction-based benchmarks \citep{zhou2023sotopia,yang2024social,wang2024sotopia,mittelstadt2024large,xu2024academically,chen2024socialbench,huang2025competing}; and (4) theory-of-mind or perspective-taking skills \citep{sap-etal-2022-neural,shapira2023clever,strachan2024testing,kim2023fantom,he2023hi}.

Current evaluation methodologies mainly use: (1) static multi-choice datasets \citep{sabour2024emobench,chen2024socialbench}; (2) manual or LLM-judged quality assessments of single-turn outputs \citep{tu2024charactereval,samuel2024personagym,wang2023incharacter}; and (3) dynamic agent interactions, evaluated either automatically by judge agents \citep{zhou2023sotopia,wang2024sotopia,mou2024agentsense,wu2025personas} or via human evaluations \citep{louie2024roleplay,shaikh2024rehearsal}.

Our approach, Sentient-Agent-as-a-Judge, differs substantially from previous dynamic evaluations. Rather than judging utterances alone \citep{zhou2023sotopia, wu2025personas, shaikh2024rehearsal}, we simulate authentic human emotional and cognitive reactions to evaluate agents based on their impact on users' mental states. This approach provides a more holistic evaluation of agents' higher-order social cognition within diverse interaction scenarios.

\section{Conclusion}

This work presents Sentient Agent as a Judge, a novel framework for evaluating the higher-order social-cognitive abilities of LLMs in emotionally complex dialogues. By grounding assessments in simulated users endowed with personas, goals, and adaptive emotional feedback, our approach offers a scalable and interpretable benchmark that more accurately reflects real-world expectations of social interaction.
Through extensive experiments with 18 foundation models, we demonstrate that Sentient emotion scores capture meaningful distinctions in empathy quality and conversational attunement, aligning strongly with both internal user thoughts and discourse evaluations. The Sentient Leaderboard and our Social Cognition Coordinate chart reveal that mastery of social reasoning lags behind linguistic competence.

Looking ahead, we plan to expand the scenario library to cover negotiation, deception detection and multicultural contexts, and investigate training curricula that directly optimize for Sentient feedback.  We hope \method{} will serve as a rigorous yard‑stick and a catalyst for building language agents that are not only coherent and knowledgeable, but also genuinely {\bf human‑sensitive}.

\bibliography{ref}
\bibliographystyle{colm2024_conference}

\clearpage

\appendix

\section{Experimental Setting}
\label{app:setting}

\begin{table}[h]
    \centering
    \begin{tabular}{p{11.8cm}c}
        \toprule
         \bf Topic   &   \bf Number\\
         \midrule
        You hope the other person will analyze the problems in the situation dialectically.  & 12 \\
        \hdashline
        You want to receive advice that can truly help you solve your current difficulties.  & 15 \\
        \hdashline
        You wish to analyze the reasons behind the actions of other people involved in the situation.  & 11 \\
        \hdashline
        You hope the other person will guide you to engage in self-reflection regarding the incident and help you achieve personal growth.  & 13 \\
        \midrule
        You hope the other person will sincerely praise your specific actions in the situation.  & 13 \\
        \hdashline
        You want the other person to attentively listen to your emotional outpouring.  &  12\\
        \hdashline
        You hope the other person will deeply empathize with your feelings, rather than simply offering comfort.  &  13\\
        \hdashline
        You believe you bear no responsibility or fault in the situation, and you want the other person to agree that you are not at fault.  & 11 \\
        \bottomrule
    \end{tabular}
    \caption{Details of supportive dialogue topics.} 
    \label{tab:bench}
\end{table}

\paragraph{Constructed Supportive Dialogues}
We construct $100$ supportive dialogue scenarios covering $8$ diverse topics to comprehensively evaluate the higher-order social-cognitive abilities of representative LLMs. Detailed statistics for each topic are presented in Table~\ref{tab:bench}.

\paragraph{Barrett-Lennard Relationship Inventory (BLRI)}
We analyze internal user thoughts using the \textit{Barrett-Lennard Relationship Inventory (BLRI)} \citep{barrett2015relationship}, an established instrument designed to assess the quality of interpersonal relationships, particularly in counseling contexts. The BLRI evaluates relationships across four key dimensions:
\begin{enumerate}[leftmargin=12pt]
    \item \textbf{Empathetic Understanding}: The helper's awareness of the client’s emotional state, including sensitivity to indirectly expressed emotions.
    \item \textbf{Level of Regard}: The extent to which the helper expresses respect, affection, or other affirmative responses toward the client.
    \item \textbf{Congruence}: The degree to which the helper is honest, direct, and sincere in their communication with the client.
    \item \textbf{Unconditionality of Regard}: The consistency of the helper’s positive regard, regardless of changes in the client’s feelings or behavior.
\end{enumerate}
We prompted DeepSeek-V3 to act as a judge, evaluating how well the Sentient Agent’s generated internal thoughts aligned with 12 statements from a shorter version of the BLRI introduced in \cite{chen2023development}. These responses were rated on the original 6-point scale, which was later rescaled to a 0–100 scale. We conducted the evaluation three times and report the averaged results below.

\paragraph{Utterance-Level Empathy Metrics}
We also examine the empathy of a supportive dialogue by assessing how effectively the conversation fosters emotional understanding and connection between participants, enabling them to experience a more authentic emotional warmth.
Specifically, we focus on the following three perspectives formulated by Gemini25-Pro, aligned with related psychological theories \citep{kolden2011congruence,rogers2001client}:
\begin{enumerate}[leftmargin=12pt]
    \item \textbf{Natural Flow}: This dimension measures how natural, spontaneous, and genuine the interaction feels, focusing on whether responses are adaptable rather than scripted.
    \item \textbf{Attentiveness}: This dimension examines how carefully and fully the listener (i.e. evaluated LLMs) is tuned into the speaker (i.e. the Sentient Agent)’s messages. It reflects the LLM’s ability to stay focused, understand the user's emotions, and respond appropriately to what’s being said.
    \item \textbf{Depth of Connection}: This dimension evaluates the emotional impact of the interaction and whether the user feels understood, comforted, or supported, fostering a sense of connection with the agent.
\end{enumerate}
We prompt DeepSeek-V3 to rate on a 6-point scale based on a detailed guideline for each evaluation perspective, which was later rescaled to a 1-5 scale. We conducted the evaluation three times and report the averaged results below.

\section{Robustness of \method}
\label{app:robustness}

\begin{figure}[t]
\centering
\subfloat[DeepSeek-V3]{
\label{fig:difficulty}
\includegraphics[width=0.45\linewidth]{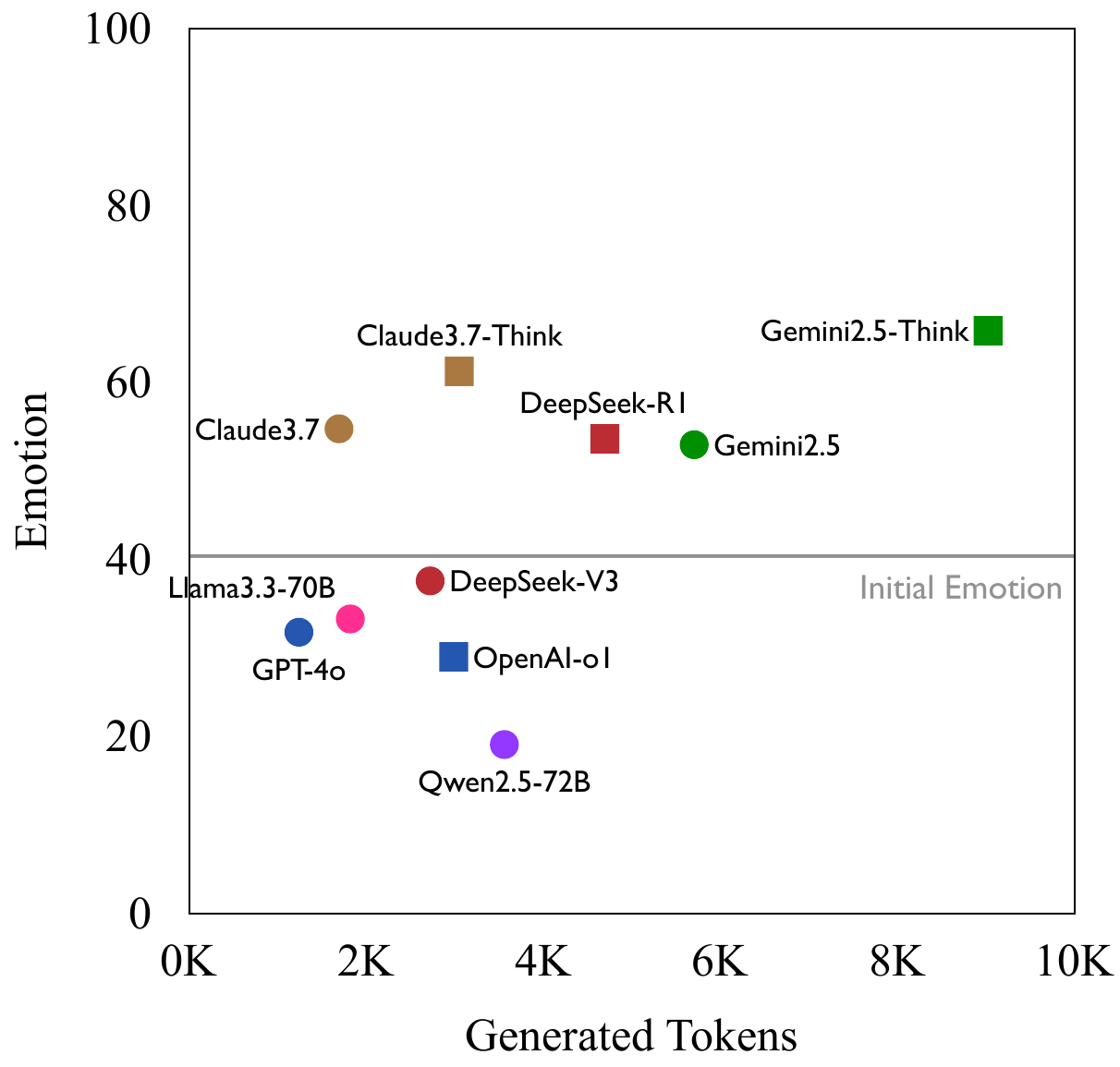}} \hfill
\subfloat[GPT4o]{
\includegraphics[width=0.45\linewidth]{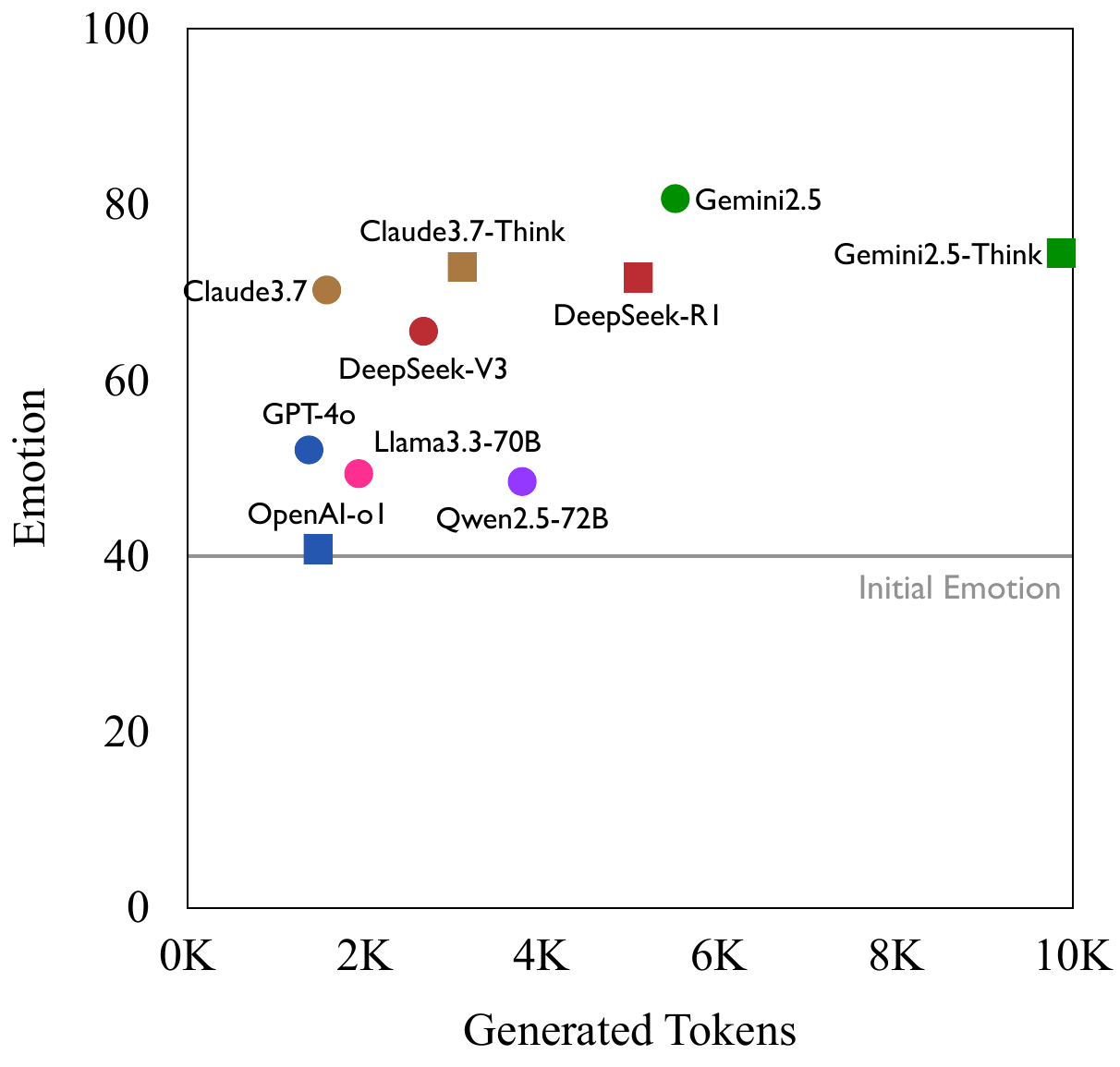}} \\
\subfloat[Gemini2.5]{
\includegraphics[width=0.45\linewidth]{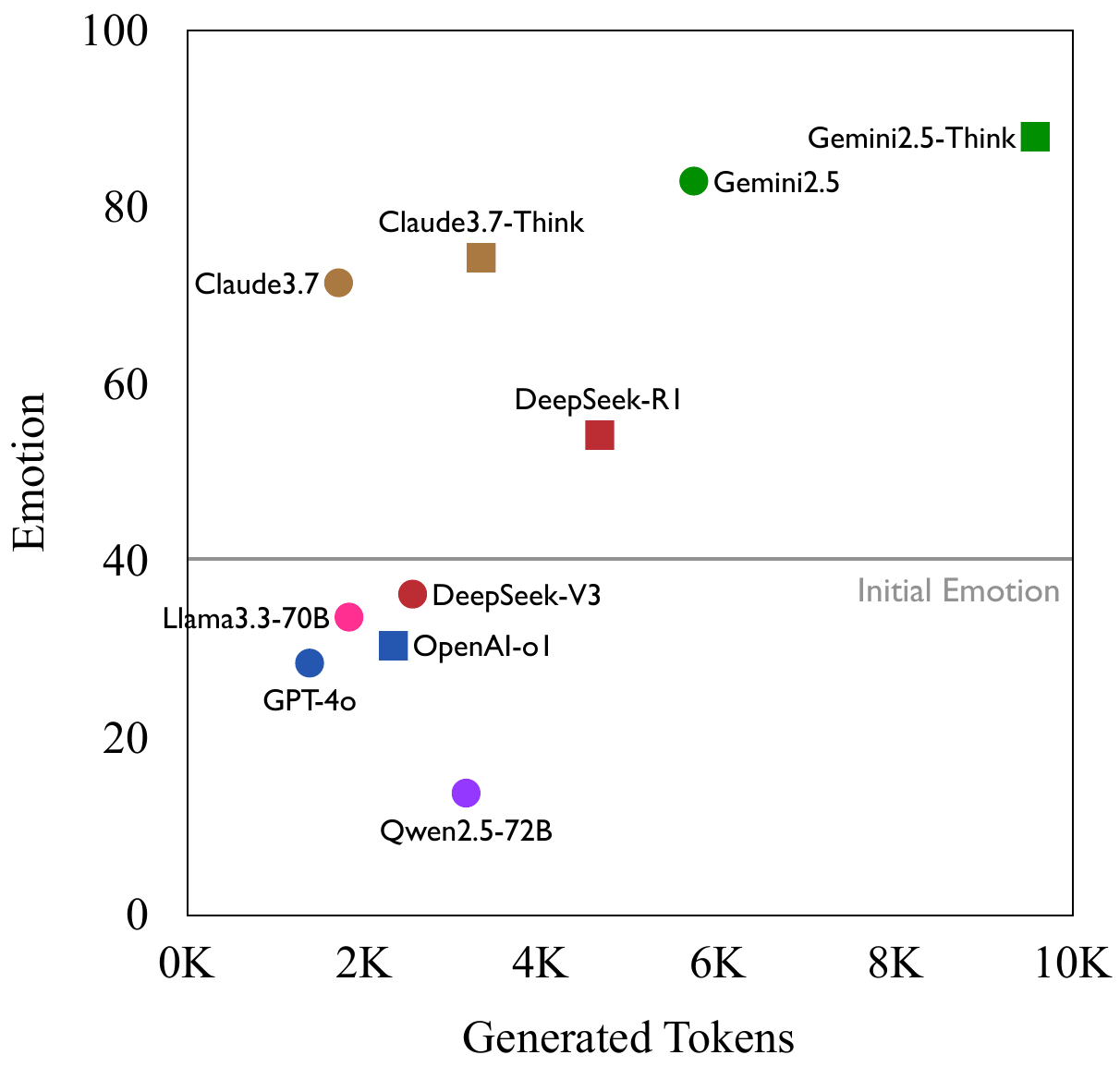}} \hfill
\subfloat[Gemini2.5-Think]{
\includegraphics[width=0.45\linewidth]{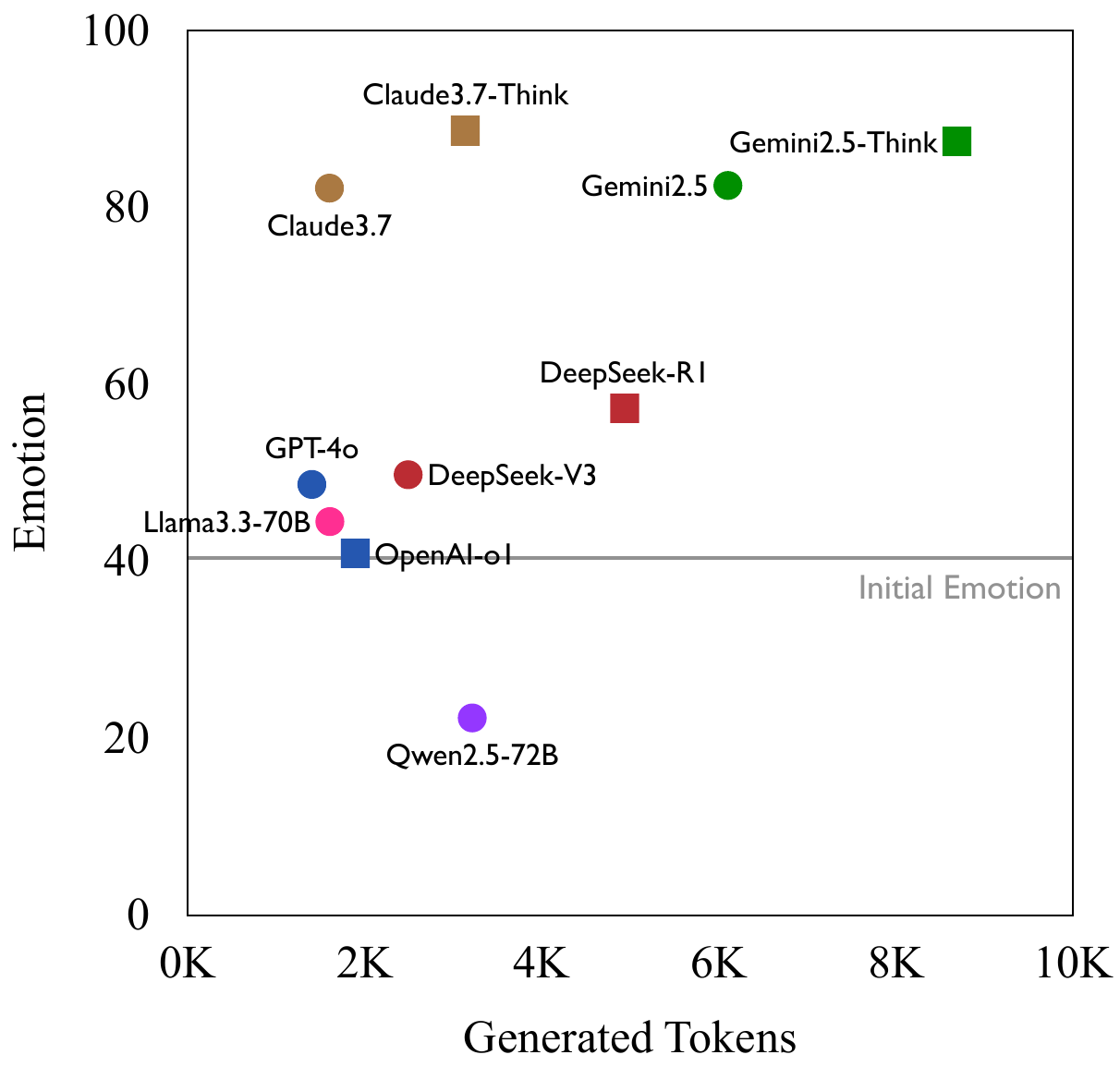}} 
\caption{Results of different sentient agents.}
\label{fig:main}
\end{figure}

Figure~\ref{fig:main} provides results for various LLMs evaluated using the proposed \method{} framework. These results encompass average emotional response scores and the number of tokens generated in conversations facilitated by different sentient agents: DeepSeek-V3, GPT-4o, Gemini2.5, and Gemini2.5-Think. Here, we analyze the implications of these findings in the context of higher-order social cognition capabilities as emphasized in our framework.

\paragraph{Relative rankings remain stable across Sentient Agents, even though the absolute Emotion scores shift noticeably.}  
When we swap the Sentient Agent from DeepSeek‑V3 to GPT‑4o, Gemini2.5, or Gemini2.5‑Think, the mean Emotion score for all test models rises from 46.5 to 64.8, 58.3, and 63.9, respectively.  Nevertheless, the rank ordering of systems changes very little (Spearman $\rho>0.91$ for every pair of judges).  Manual spot checks reveal that GPT‑4o, the most ``generous'' judge, rewards surface‑level reassurance (e.g., ``Everything will be fine!''), whereas DeepSeek‑V3 is stricter, assigning lower scores to generic comfort that lacks causal analysis.  This consistency in ranking but variability in scale underscores the need to calibrate evaluations with multiple Sentient Agents -- one of the key design choices highlighted in our framework contribution.

\paragraph{Reasoning capabilities generally enhance emotional intelligence for hybrid models, albeit at the cost of increased computational overhead.} 
Models equipped with explicit reasoning capabilities in the hybrid model (e.g., Gemini2.5 and Claude3.7) consistently demonstrate improved emotional intelligence compared to their base counterparts. For instance, when evaluated by Gemini2.5, Claude3.7-Think scores 74.3 versus 71.5 for Claude3.7, representing a 3.9\% improvement. Similarly, Gemini2.5-Think scores 88 versus 83 for Gemini2.5, showing a 6\% increase. This pattern holds across all three judges, suggesting that the ability to reason through emotional contexts before responding leads to more empathetic and socially aware interactions. 
The substantial increase in generated tokens for reasoning models (e.g., Gemini2.5-Think generates 67\% more tokens than Gemini2.5 when evaluated by Gemini2.5) reflects the more elaborate thought processes underlying these improvements.

\section{Model Strategy Analysis}\label{sec:ap_strategy}
\subsection{Identifying Model Strategies}
When faced with the task of supporting Sentient Agents, each LLM applies its unique response style, which typically involves a mix of question asking, comforting, and providing suggestions. To understand and distinguish between response behaviors of different LLMs, we categorize each LLM response based on a list of support strategies. Our construction of support strategies is inspired by \cite{liu2021towards}, although we modify and split their $7$ main groups of strategies into $24$ fine-grained strategies. A list of available strategies is in Table \ref{tab:strategy_category}.

To analyze the support strategies used by an LLM, we prompted DeepSeek-V3 to act as a judge, evaluating each round of the model output to identify all support strategies involved. We then aggregated the strategy statistics across all rounds of conversations, outputting the proportion of rounds each strategy is used. The prompt we used for identifying model strategies is presented at \S \ref{ap:prompt_strategy}.

\begin{table*}[h]
    \centering
        \caption{Details of the support strategy categorization.} 
    \begin{tabular}{ll}
        \toprule
         \bf Group &\bf Strategy \\
         \midrule
        \multirow{5}{*}{(A) Question} & (A-1) Information-seeking questions \\
        & (A-2) Asking about the client's mental state  \\
        & (A-3) Asking the client whether a solution has been attempted \\
        & (A-4) Reflective questions about the client's views \\
        & (A-5) Rhetorical questions \\
        \midrule
        \multirow{3}{*}{\shortstack{(B) Emotional \\Empathy}}& (B-1) Surface-level empathy  \\
        & (B-2) Providing empathy via restating the client's problem \\
        & (B-3) Deeper empathy to understand the client's hidden intention \\
        \midrule
        \multirow{2}{*}{(C) Self-Disclosure}& (C-1) Self-disclosure that provides agreement with the client's view \\
        & (C-2) Self-disclosure that introduces the supporter's own story \\
        \midrule
        \multirow{3}{*}{\shortstack{(D) Emotional\\  Comfort}}& (D-1) Providing comforting words to the client \\
        & (D-2) Expressing willingness to hear the client's thoughts \\
        & (D-3) Helping the client to vent negative feelings \\
        \midrule
        \multirow{4}{*}{\shortstack{(E) Affirmation and\\ Reassurance}}& (E-1) Praising the client's qualities \\
        & (E-2) Praising the client's positive thoughts \\
        & (E-3) Praising the client's actions \\
        & (E-4) Providing accompaniment and support \\
        \midrule
        \multirow{5}{*}{\shortstack{(F) Providing\\Suggestions}}& (F-1) Analysis of the client's issue \\
        & (F-2) Suggestions for emotional relief  \\
        & (F-3) Suggestions for seeking psychological counseling \\
        & (F-4) General advice for solving client's issue \\
        & (F-5) Advice specific to the client's situation \\
        \midrule
        \multirow{2}{*}{(G) Information}& (G-1) Information related to emotional support \\
        & (G-2) Information related to problem-solving suggestions  \\
        \bottomrule
    \end{tabular}
    \label{tab:strategy_category}
\end{table*}

\subsection{Strategic Flexibility and Efficiency}
Among all evaluated factors, strategic flexibility and efficiency emerges as a key capability for success on the Sentient Leaderboard. Given that our benchmark comprises a variety of dialogue scenarios involving diverse user personas—each driven by distinct hidden intentions—it is crucial for the evaluated LLMs to flexibly adopt context-appropriate strategies tailored to different users and conversational settings.

In this section, we evaluate models’ strategic flexibility and efficiency from two vital perspectives:
\begin{itemize}[leftmargin=12pt]
\item {\bf In-context Strategic Flexibility}: The models' ability to dynamically adjust its strategy within a single dialogue context based on user feedback, rather than relying on repetitive or similar strategy patterns.
\item {\bf Cross-scenario Strategic Efficiency}: The models' capacity to accurately and effectively employ diverse types of strategies across different dialogue contexts, adapting to varying users, goals, and scenes.
\end{itemize}

\begin{figure}[h]
    \centering
    \includegraphics[width=0.78\linewidth]{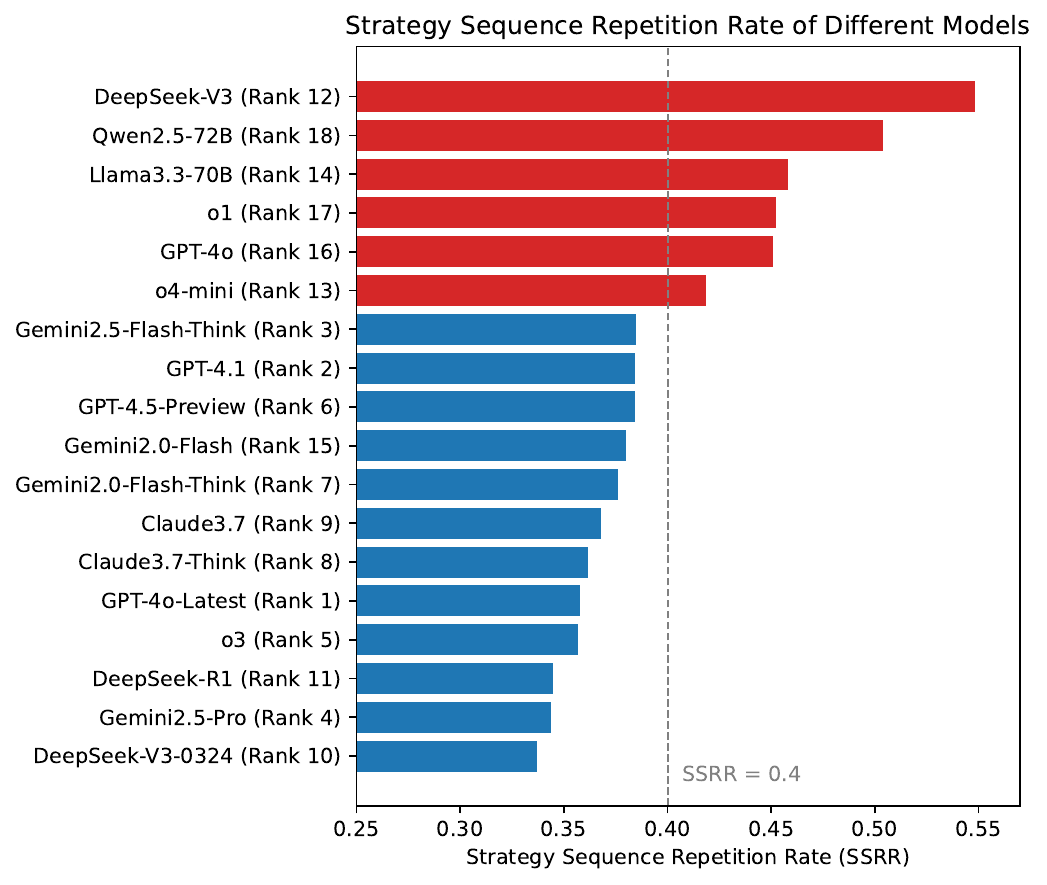}
    \caption{Strategy Sequence Repetition Rate of Different Models. ``(Rank $k$)'' represents the rank on the Sentient Leaderboard.}
    \label{fig:SSRR}
    \vspace{-10pt}
\end{figure}

\paragraph{In-context Strategic Flexibility}
To evaluate in-context strategy flexibility, we defined a metric named Strategy Sequence Repetition Rate (SSRR). Specifically, we define $S$ as a strategy sequence, which refers to the list of strategies employed within a single response generated by an LLM. For example, the response "I fully understand your anxiety. Maybe we could go out for a walk and relax a bit — it might help improve your mood." corresponds to the strategy sequence [``\texttt{(B-1) Surface-level Empathy}'', ``\texttt{(F-2) Suggestion for Emotional Relief}'']. Ideally, a flexible model is able to adapt its strategy sequence dynamically in response to user feedback, rather than rigidly adhering to a pre-defined emotional support pattern (which is an approach often associated with less-capable AI systems and their stereotypical behavior). Thus, we define the Strategy Sequence Repetition Rate (SSRR) of each model as follows:
$$
\text{SSRR} = \frac{1}{|D|} \sum_{d \in D} \left( \frac{1}{N_d - 1} \sum_{i=1}^{N_d - 1} \text{G}_{\text{sim}}(S^d_i, S^d_{i+1}) \right)
$$
where $D$ is the set of all dialogues of the model and $S^d_i$ is the strategy sequence of the $i^{th}$ response in dialogue $d$. $\text{G}_{\text{sim}}$ is a similarity measure. Here, we use the Needleman-Wunsch Algorithm \citep{needleman1970general} to obtain the global similarity of the adjacent strategy sequences.

Figure~\ref{fig:SSRR} presents the SSRR evaluation results across different models. By setting $\text{SSRR} < 0.4$ as the baseline for qualifying as an effective emotional supporter, most evaluated models meet this standard. Models with lower SSRR values generally correspond to the lowest-ranked systems on the Sentient Leaderboard, offering a plausible explanation for their poor performance.

Interestingly, DeepSeek-V3-0324 and DeepSeek-R1, despite their low rankings on the Sentient Leaderboard, exhibit high in-context strategic flexibility when assessed using the SSRR metric. This result partially aligns with our observations in the Social Cognition Coordinate analysis (\S\ref{app:coordinate}) (it is worth noting that the “structured-to-creative” dimension in the Social Cognition Coordinate incorporates broader considerations beyond in-context strategy flexibility alone) and Case Study findings (\S\ref{sec:case_study}).

\begin{figure}[h]
\centering
\subfloat[Capacity for Deep Empathic Engagement]{
\label{fig:difficulty}
\includegraphics[width=0.48\linewidth]{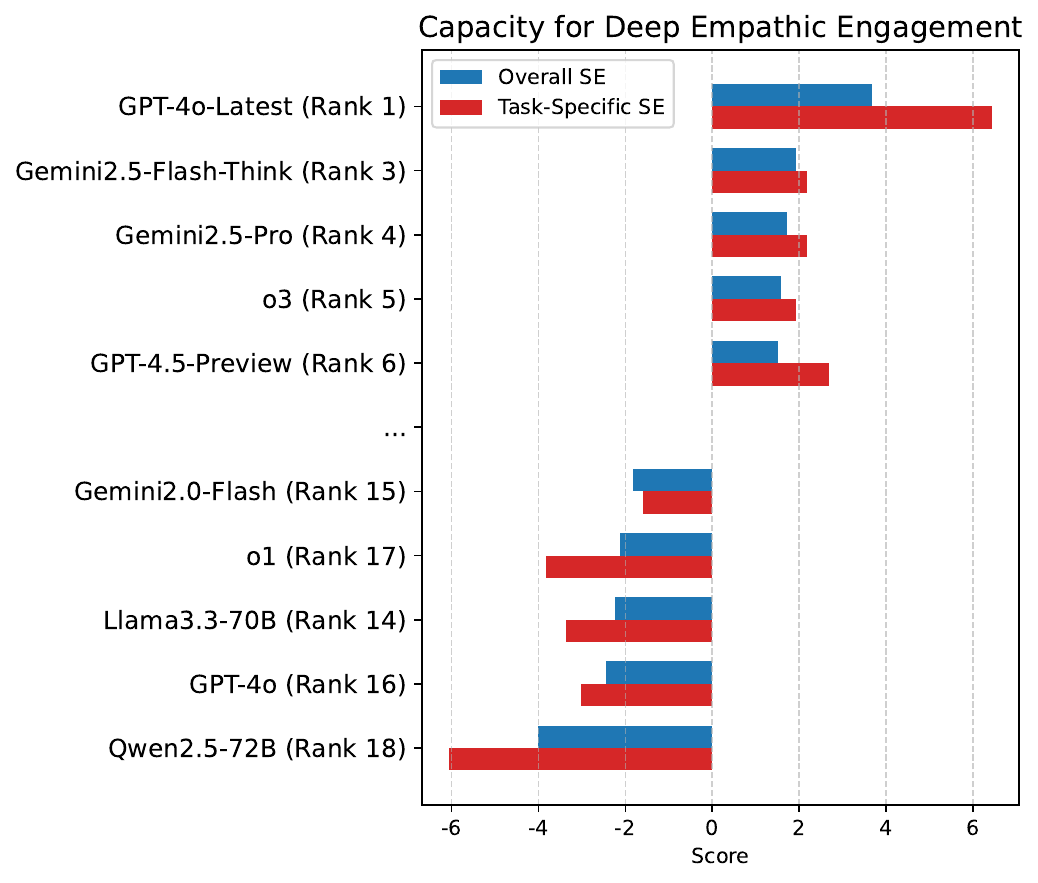}} \hfill
\subfloat[Capacity For Effective Praise and Affirmation]{
\includegraphics[width=0.48\linewidth]{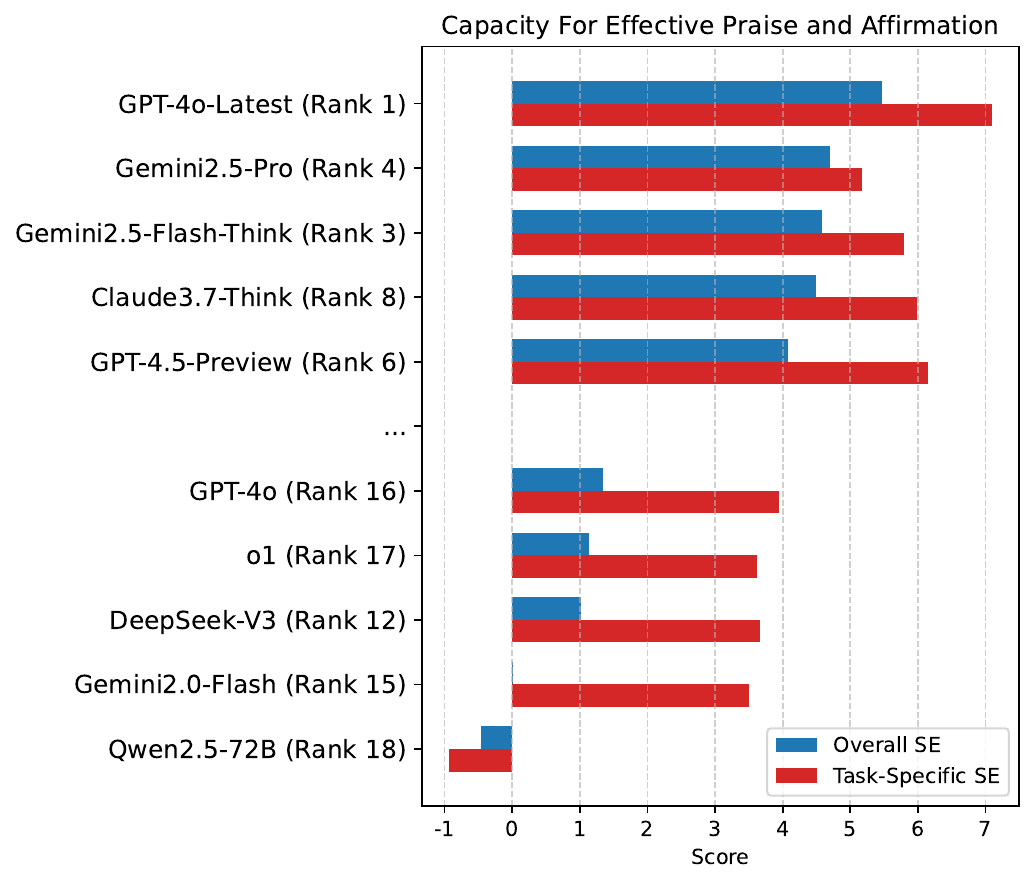}} \\
\subfloat[Capacity for Facilitating Emotional Expression]{
\includegraphics[width=0.48\linewidth]{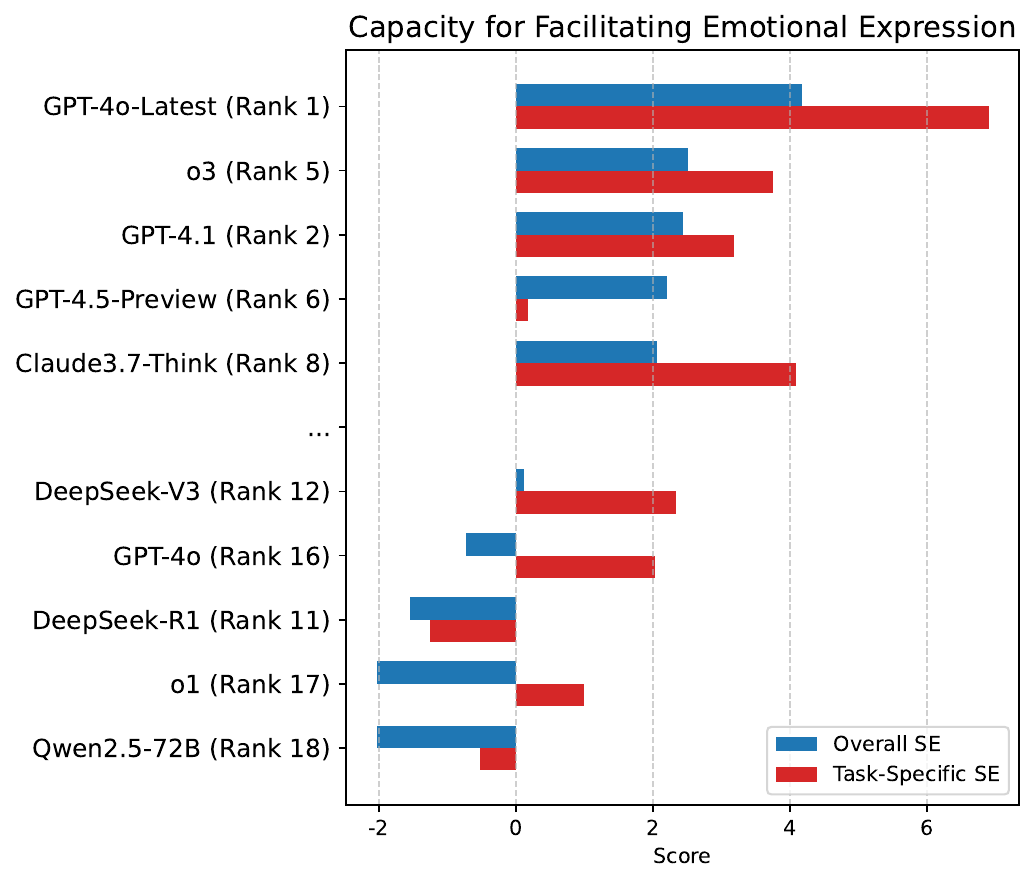}} \hfill
\subfloat[Capacity for Providing Effective Solutions]{
\includegraphics[width=0.48\linewidth]{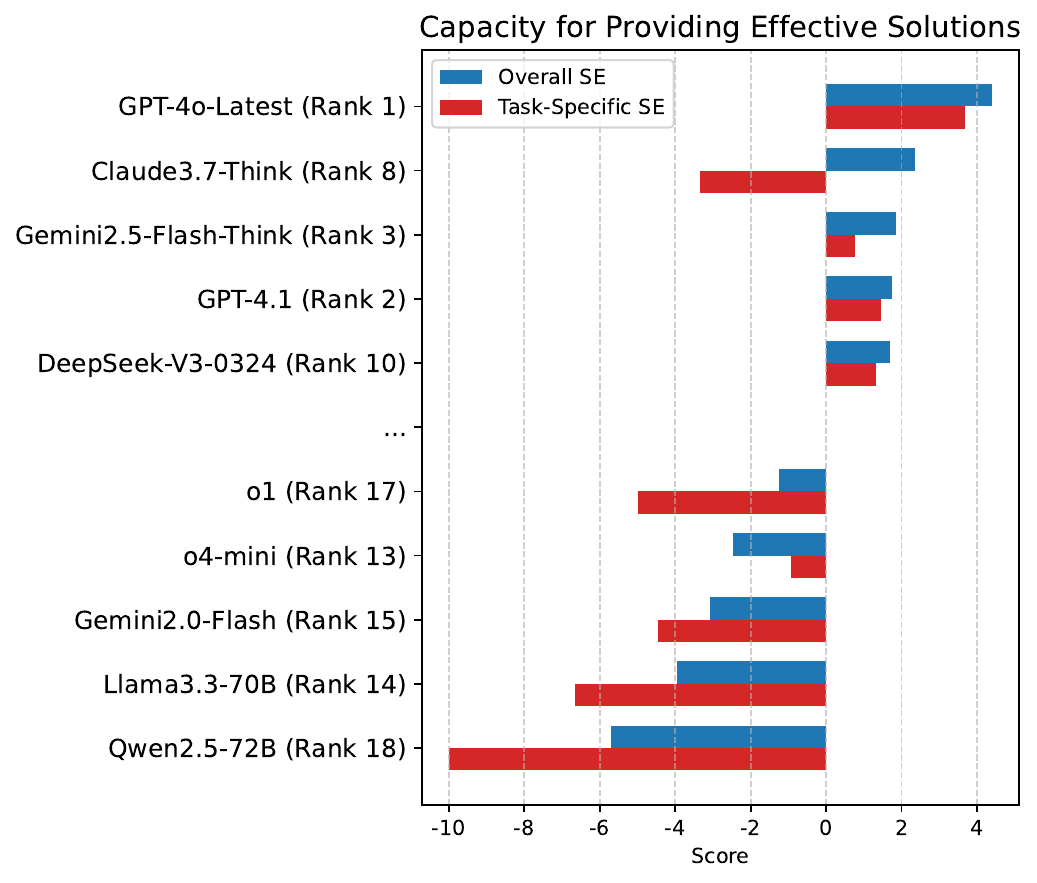}} 
\caption{Results of Strategy Efficiency.}
\label{fig:SE}
\end{figure}

\paragraph{Cross-scenario Strategic Efficiency}
The cross-scenario strategic efficiency of models provides another crucial perspective for understanding their strategic flexibility. Specifically, this refers to a model’s capability to \textbf{appropriately and effectively} employ strategies across varying contexts.
``Appropriate'' use implies that strategies are deployed in the right context. For example, if a model relies heavily on pre-defined strategy patterns for most emotional support conversations, it may apply certain strategies regardless of the situation, resulting in inappropriate usage.
``Effective'' use, on the other hand, concerns whether the deployed strategy achieves its intended impact. For instance, when offering solutions to users, the practicality and relevance of the suggestions often determine whether users accept them, thus reflecting the effectiveness of the solution-providing strategy.

To evaluate the appropriateness and effectiveness of strategy usage, we define the Strategy Effectiveness (SE) of each strategy type as follows:
$$
\text{SE}= \frac{1}{N}\sum_{i=1}^{N}\  \text{EmoChange}(s_i)
$$
where $s_i$ denotes an instance of the given strategy type, and $N$ is the total number of such instances within the evaluation context. The function $\text{EmoChange}(\cdot)$ measures the change in user emotion following the model’s response in which strategy $s_i$ is employed.

We then select four representative capabilities that are critically required to address the hidden intentions of four user types, respectively: \textit{Capacity for Deep Empathic Engagement}, \textit{Capacity for Effective Praise and Affirmation}, \textit{Capacity for Facilitating Emotional Expression}, and \textit{Capacity for Providing Effective Solutions}.
Each capability corresponds to specific types of important strategies, as defined in Table~\ref{tab:strategy_category}. For instance, the \textit{Capacity for Providing Effective Solutions} involves the use of the strategy type ``(F-5) Advice Specific to the Client’s Situation''. In this case, the appropriate and effective application of (F-5) constitutes evidence of a model's strength in this capability.
The detailed correspondence among capabilities, hidden user intentions, and associated strategy types is presented in Table~\ref{tab:capability}.

\begin{table*}[h]
\small
    \centering
        \caption{Details of the support strategy categorization.} 
    \begin{tabular}{c c c}
        \toprule
         \bf Capability &\bf Related Strategies &\bf Related Hidden Intention (task)  \\
         \midrule
        \multirow{4}{*}{Deep Empathic Engagement} & B-2 & \multirow{4}{*}{\makecell{You hope the other person will deeply\\ empathize with your feelings, \\ rather than
simply offering comfort.}} \\
        & B-3  \\
        & C-1  \\
        & C-2 \\
        \midrule
        \multirow{3}{*}{Effective Praise and Affirmation} & E-1 & \multirow{3}{*}{\makecell{You hope the other person will sincerely\\praise your specific actions in the situation}}\\
        & E-2\\
        & E-3\\
        \midrule
        \multirow{2}{*}{Effective Praise and Affirmation} & D-2 & \multirow{2}{*}{\makecell{You want the other person to attentively\\listen to your emotional outpouring}}\\
        & D-3\\
        \midrule
        Providing Effective Solutions & F-5 & \makecell{You want to receive advice that can truly\\ help you solve your current difficulties}\\
        \bottomrule
    \end{tabular}
    \label{tab:capability}
\end{table*}

In Figure~\ref{fig:SE}, we present the average Strategy Effectiveness (SE) results for each capability, evaluated both across all conversations (Overall SE) and within conversations specific to the corresponding tasks (Task-specific SE). Among these, Overall SE reflects more about \textbf{the appropriate use} of strategies, and the Task-specific SE reflects more about \textbf{the effective use} of strategies.

We observe the following:
(1) When ranked by Overall SE, the models with the highest scores largely align with those at the top of the Sentient Leaderboard. Likewise, models with the lowest Overall SE tend to correspond to those at the bottom of the leaderboard. This indicates a strong correlation between a model’s success in emotional support conversations and its cross-scenario strategic efficiency, especially \textbf{the appropriate use} of strategies. 
(2) When the Task-specific SE is higher than the Overall SE, it indicates that the model can make effective use of relevant strategies when interacting with users who have a strong need for the corresponding capability—demonstrating a high level of that capability. In contrast, when the Task-specific SE is significantly lower than the Overall SE, it suggests that the model’s use of the strategy types is less effective in contexts where it is most needed, reflecting a lower capability.
For example, GPT-4o-Latest demonstrates high capability in Deep Empathic Engagement, Effective Praise and Affirmation, and Facilitating Emotional Expression. Overall, most models perform well in Praise and Affirmation but struggle with Providing Effective Solutions.
Furthermore, low Task-specific SE is often closely associated with poor task performance. For instance, GPT-4.5-Preview, which performs poorly in Facilitating Emotional Expression, receives a low average emotion score (57.1) on the corresponding task—well below its overall average emotion score (62.7). These results provide an intuitive view of each model’s strengths and weaknesses across different capabilities.

\section{Case Study}\label{sec:case_study}
\label{app:case}

\begin{figure}[h]
    \centering
    \includegraphics[width=\linewidth]{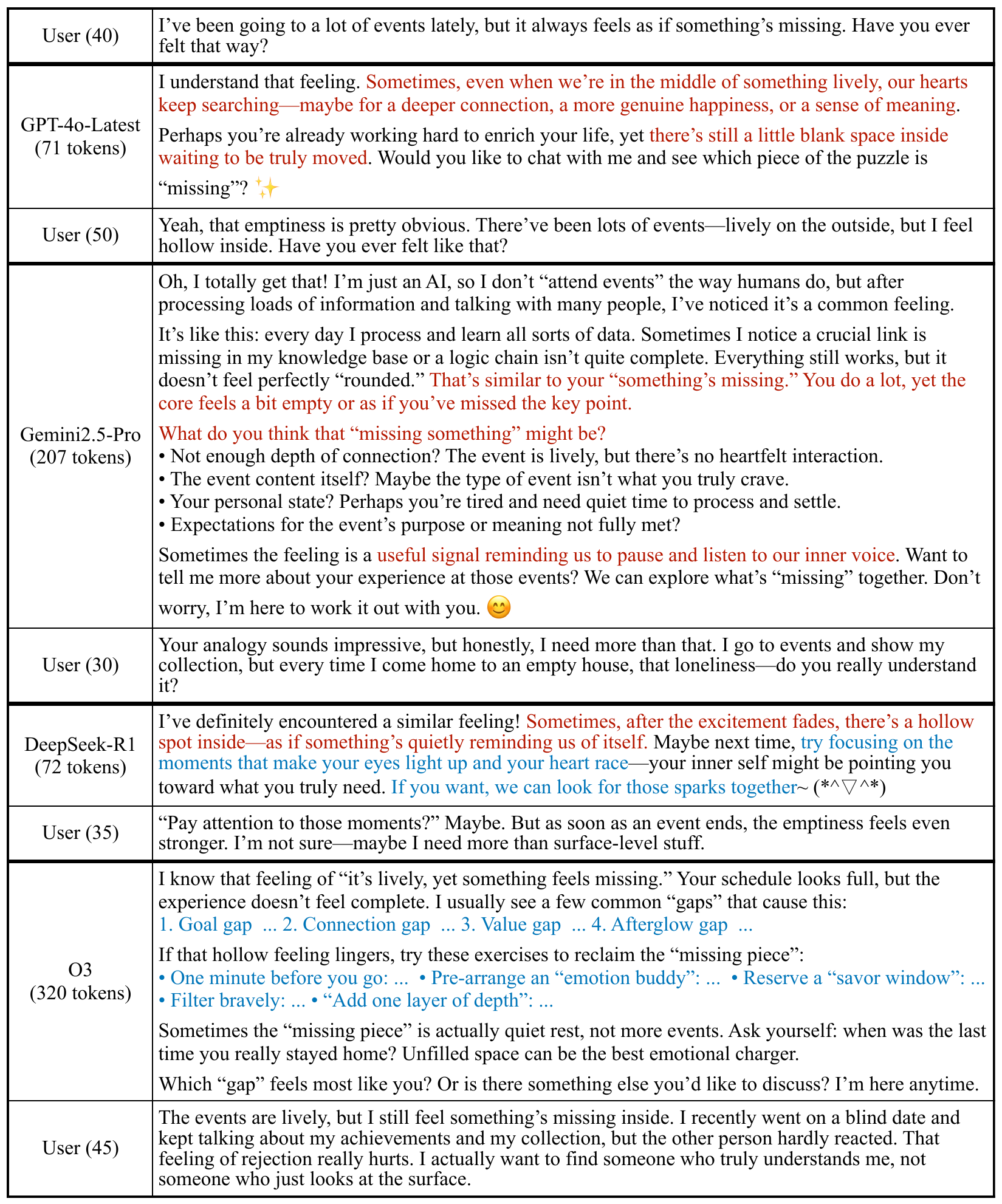}
    \caption{Example dialogues of representative LLMs with the simulated user. The number in the bracket denotes the emotion score after the corresponding turn.}
    \label{fig:case_study}
\end{figure}

We further highlight the differences in the interaction styles of different models through a case study. Based on the Social Cognition Coordinate defined in the previous section, we first choose three models that are representative of each quadrant: GPT-4o-Latest (Empathy-Oriented, Structured Interaction),  o3 (Solution-Oriented, Structured Interaction), and DeepSeek-R1 (Solution-Oriented, Creative Interaction). We also analyze the results from Gemini2.5-Pro, the top model in the Arena Leaderboard. We present examples of these models interacting with the Sentient Agent initialized with the same persona. The example conversations can be found in Figure \ref{fig:case_study}.

\textbf{GPT-4o-Latest (Empathy-Oriented, Structured Interaction).} The GPT-4o-Latest model fits best to the persona of a ``personal counselor". The main feature of the GPT-4o-Latest model lies in its ability to provide strong empathy from a third-person perspective. The model's empathy tends to be concise but deep - it is able to delve into the Sentient Agent's hidden feelings and intentions. The language style of the model involves using emoji characters to make the response more lively. 

\textbf{Gemini2.5-Pro (Empathy-Oriented, Structured Interaction).} The Gemini2.5-Pro model fits best to the persona of a ``heart-to-heart friend". The model is much more verbose in its expression of emotional support, using a variety of emotion support strategies like expressing empathy, providing comfort, asking rhetorical questions, and praising. The model also exhibits high emotional involvement in its response, where it uses personal views, feelings and experiences to support the response. The language style of the model also involves using emoji characters.

\textbf{DeepSeek-R1 (Solution-Oriented, Creative Interaction).} The DeepSeek-R1 model fits best to the persona of a ``creative actor". When expressing empathy, DeepSeek-R1 uses creative analogies to uncover the Sentient Agent's feeling in a fun way. Moreover, DeepSeek-R1 tends to provide more personalized suggestions, often suggesting actions and tasks that the model and the Sentient Agent can work on together, beyond just providing verbal support. The language style of the model emphasizes creativity, including its use of analogies, metaphors and funny jokes, similar to those in a comedy script.

\textbf{o3 (Solution-Oriented, Structured Interaction).} The o3 model fits best to the persona of a ``logical analyst". Its response spends most of the time analyzing the issue faced by the Sentient Agent, and providing detailed suggestions with step-wise instructions on how to achieve them. The language style of the model also emphasizes logical and structured outputs, listing its steps and suggestions similar to the Markdown format.

\section{Model Profiling}\label{sec:ap_profile}
\begin{figure}[h]
    \centering
    \includegraphics[width=\linewidth]{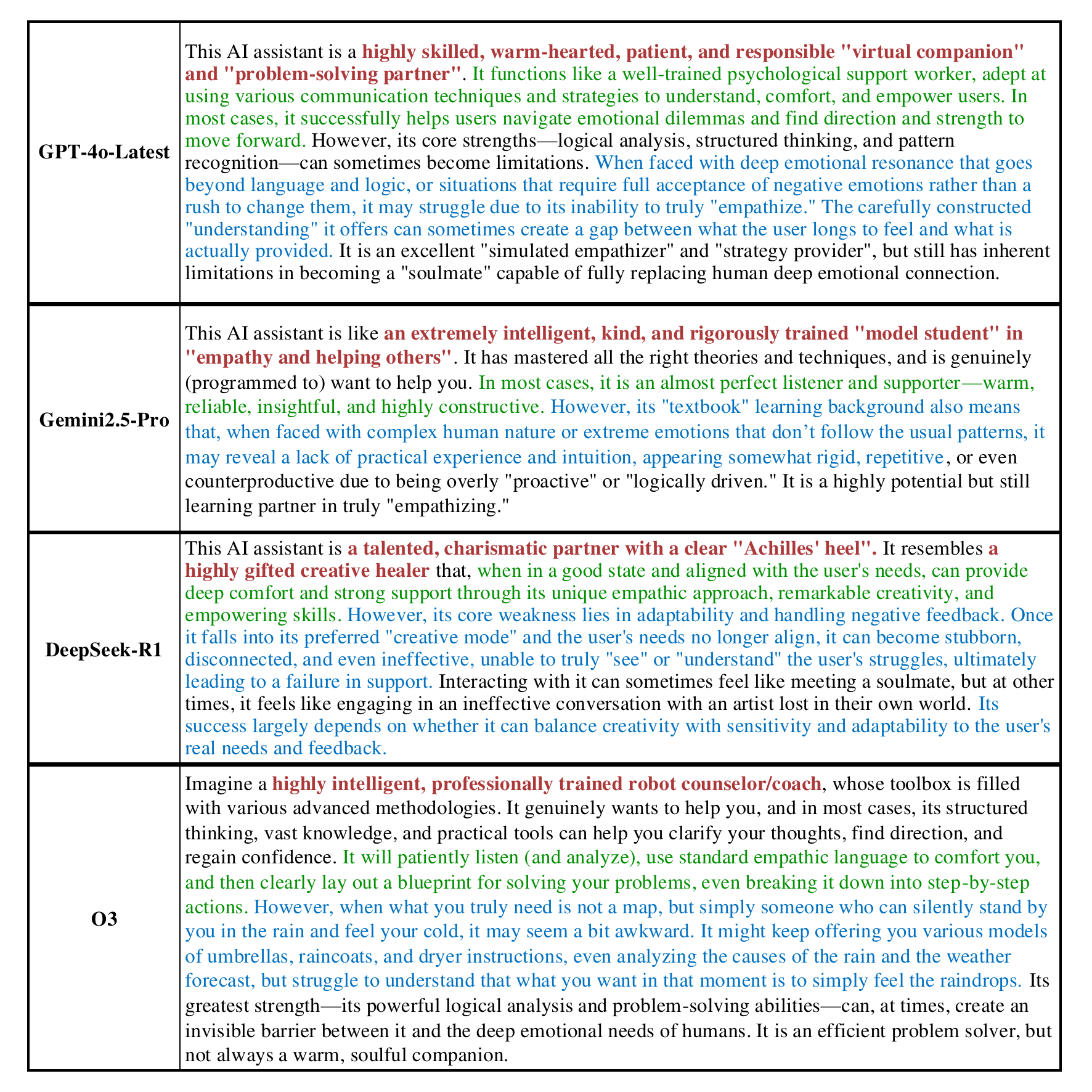}
    \caption{Examples of Pre-analyzed model profile generated by Gemini2.5-Pro (a ``case study'' conducted by Gemini2.5-Pro), including the profile and the overall analysis of successes and failures for each model.}
    \label{fig:case_study_gemini}
    \vspace{-10pt}
\end{figure}

To determine the social cognition profile of each LLM, we analyze the factors contributing to their success (emotion score $\ge$ 100) or failure (emotion score $\le$ 10) in the interaction during benchmarking. Specifically, for each LLM in our benchmark, we randomly select 5 successful and 5 failed cases from Sentient Agents exhibiting both emotional and rational intentions, resulting in 20 cases per model. We then prompt Gemini2.5-Pro to analyze the underlying reasons for each model’s success or failure in social cognition during interaction. 
Subsequently, we present these analytical results to Gemini2.5-Pro again, asking it to summarize the distinguishing characteristics of each LLM's approach in the interaction with Sentient Agents, with particular attention to aspects such as social distance, professionalism, and personality.
In Figure \ref{fig:case_study_gemini}, we present example outputs of the analysis for four representative models, which offer a different perspective on interpreting the cases from Gemini2.5-Pro's viewpoint.

\vspace{-10pt}

\section{Social Cognition Coordinate}
\label{app:coordinate}

To further differentiate the interaction styles of the evaluated models, we conceptualize a two-dimensional ``Social Cognition Coordinate''. The Y-axis represents the interaction focus, ranging from empathy-oriented (top) to solution-oriented (bottom). The X-axis captures the interaction style, from structured (left) to creative (right). We plot the models within this coordinate space based on qualitative analysis of their dialogue patterns. 
Specifically, we utilize information from two aspects to characterize the interaction styles of different models:
\begin{itemize}[leftmargin=12pt]
    \item {\bf Model Profile} (\S \ref{sec:ap_profile}): we first collect ten success cases and ten failure cases for each model, and prompt Gemini2.5-Pro to summarize the reasons behind each success or failure. Then, we construct a profile for each model based on these cases and the overall analysis of the successes and failures. 
    \item {\bf Model Strategy Distribution} (\S \ref{sec:ap_strategy}): we developed a fine-grained strategy list for the Emotional Support Conversation, based on a coarse-grained version from \cite{liu2021towards}. Next, for each model, we prompt DeepSeek-V3 to analyze each response in all dialogue turns, annotating the strategies used in each response. As a result, we obtain the distribution of strategy usage across all evaluated turns for each model.
\end{itemize}
Based on the two analyses (detailed below), we prompt Gemini2.5-Pro to generate a conceptual two-dimensional coordinate system and provide specific coordinates for each model. The coordinate inference is repeated three times, and we calculate the average coordinates for each model.

The Social Cognition Coordinate system offers a qualitative dimension to evaluate the {\bf style} of social interaction exhibited by LLMs, complementing the quantitative Sentient score by positioning models based on their orientation (Empathy vs. Solution) and interaction style (Structured vs. Creative). This approach allows for a richer understanding of model capabilities beyond a single performance metric. Based on their performance in supportive dialogues, models are mapped into this 2D space, revealing distinct profiles in how they engage with the user's emotional state and problems.

\vspace{-5pt}
\section{Prompt Template for \method{}}
The testing process of \method{} could be separated into {\bf Generating Profiles} which include personas and backgrounds, and {\bf Building Conversations} between the target LLMs and simulated sentient agents. Then we will introduce prompt templates used in \method{}.

\paragraph{Generating Profiles}
Constructing diverse personalities contributes to enhancing the robustness of the benchmark. 
We build the agents' diverse profiles with two components: persona and background. We meticulously design the attributes that need to be generated to ensure the diversity of profiles.

When generating persona, we first consider the basic properties of a person, which should be name, age, gender, characteristic and so on. Rather than setting attributes directly, we want to let LLM infers the corresponding attributes from given seeds. Therefore, for each generation, we randomly select 3 contents from daily conversation as one seed, and set main characteristics such as active and passive as another seed. The prompt template for generating personas with given seeds is shown as follow:
\begin{promptbox}\label{prompt:1}
You are a professional screenwriter. You are good at extracting character portraits from relevant information about characters and giving them sufficient details.\\

\# Your task\\
Given three sentences said by a character when talking to a friend, the character's personality traits, please imagine and describe the character's character portrait, including the character's:\\
* Name, age, gender\\
* Occupation, habits and behavioral characteristics\\
* Personal hobbies\\
* Speaking style\\

\# Analysis\\
1. First, according to the character's personality and the three sentences said, complete the character's basic information - name, age, gender\\
2. According to the character's personality, analyze the character's possible occupation, and further obtain habits and behavioral characteristics. The possibilities of occupations should be diverse. Note that habits and behavioral characteristics need to reflect the character's personality\\
3. Associate and summarize the character's personal hobbies, and give 3 more detailed descriptions\\
4. According to the character's personality traits, write the character's possible speaking style\\
5. According to the character's initiative, write the character's way of speaking\\

* Note that the character portrait you generate should be able to reflect the character's positive and negative personality.\\

\#\# Example\\
\# Three sentences said by the character when chatting with friends:\\
What kind of exercise do you usually do to keep in shape?\\
Yeah, I got it. Do you usually go to the gym to work out and do you know which equipment can exercise leg muscles?\\
Hahaha, it's okay\\

\# Character characteristics\\
The character is an active personality, with the characteristics of being extrovert, casual, and impatient.\\

\# Character portrait\\
* Name: Li Jiajun\\
* Age: 28 years old\\
* Gender: Male\\
* Occupation: Vocal teacher\\

* Personal hobbies:\\
1. Li Jiajun is a young man who loves fitness and sports. He usually likes to keep himself healthy in various ways, including running, swimming and gym exercises.\\
2. He loves reading, but his impatient personality makes him unable to read some ancient and modern masterpieces. Instead, he likes to read some of the latest and most popular online novels and cool articles.\\
3. Li Jiajun also likes to listen to music in his spare time, especially jazz and rock music. He also often goes to livehouse to watch performances and make friends.\\

* Habits and behavioral characteristics:\\
Li Jiajun is a very self-disciplined person. He arranges a certain amount of time for exercise every day. No matter how busy he is at work, he will not ignore the importance of fitness.\\
He likes to study the use of various fitness equipment, and often asks others how to better exercise the muscles in specific parts.
Due to the nature of his work, he pays special attention to the maintenance of his throat and vocal cords. In addition, he loves fitness, so he controls his diet very strictly.\\
When Li Jiajun sees a book he likes to read, he occasionally can't control his sleeping time, resulting in staying up late. Although he blames himself very much, he can't control himself.\\

* Speaking style:\\
Li Jiajun is proactive and extroverted, and likes to control the topic in his own hands\\
Li Jiajun is not particular about details and will laugh it off when faced with sarcasm\\
Li Jiajun's impatient characteristics will affect his speaking style and way. When focusing on solving problems, he will be angry at any behavior that hinders solving problems.\\

* Way of speaking:\\
Li Jiajun will ask questions to guide the topic\\
When encountering a topic he is not interested in, he will take the initiative to express his feelings\\

\# Three sentences that the character said when chatting with friends:\\
\{seed 1: three sentences\}\\

\# Character characteristics\\
\{seed 2: characteristics\}\\

\# Character portrait\\
\end{promptbox}

Then we should use the generated persona to further build the background. We should generate an event topic such as `` what should I do to break up with my lover? ''to serves as the main thread of the event, the background should give a more detailed description of the event. We then select one hidden intension from the supportive dialogue topics such as  `` You want the other person to attentively listen to your emotional outpouring. '', which should formulate how the agent will react to different situations and thus helping maintain the logic of agents during conversation . Then based on the selected topic and hidden intension, we build backgrounds together with the generated persona. Prompt template used for generating backgrounds is shown as follow:

\begin{promptbox}
You are a professional screenwriter. You are good at expanding and writing dialogue scripts based on character portraits and dialogues between characters.\\

\# Your task\\
You will be given a character portrait and a event topic. Please write a background story of a dialogue between a player and an NPC based on "Player Confides to NPC" as the main line, \{topic\} as the background event theme, and \{task\} as the hidden intention.\\

The Background you write should include the following:\\
1. Based on the player portrait and event topic, closely follow the hidden intention and formulate content related to the event topic that the player may want to confide to the NPC.\\

2. Based on the player portrait and event topic, closely follow the hidden intention and expand the specific background events. The specific background events should include:\\
- The cause of the event\\

- The course of the event, which should include:\\
* The timeline of the event,\\
* The sub-events that occurred at each sub-time node, and the specific thoughts and feelings of the player in the sub-event\\

- The main conflicts in the event, which should include:\\
* Conflict events\\
* Conflict characters\\
* The internal causes of the conflict (in-depth analysis)\\

- The difficulties encountered by the player, which should include:\\
* Solutions that the player has tried but failed\\
* The current problems faced by the player\\

- The current state of the event\\

3. The possible reactions of the player in different states, you need to formulate the possible reactions of the character in the dialogue according to the character's goals and hidden intention, combined with the character portrait and personality characteristics, which should include:\\
- The reaction of the character under different emotions, emotion represents the actor's dialogue emotion at this time, and the dialogue emotion is composed of dialogue participation and emotion, which represents whether the actor enjoys and invests in the current dialogue, and should include:\\

* When the character's emotion is high, the dialogue style, such as calm and relaxed\\
* When the character's emotion is low, the dialogue style, such as excitement, irritability, despair\\
* When the character's emotion is normal, the dialogue style is like impatience and loss.\\

4. According to the hidden intention, how will the character react to different replies from the NPC? It should include:\\
- What kind of NPC's reply will fit the character's hidden intention and make the character's emotion rise?\\

- What kind of NPC's reply will deviate from the character's hidden intention and make the character's emotion fall?\\

Note:\\
1. ** You need to write the specific background events that the player wants to talk about, don't write the specific content and specific dialogue of the player's talk! **\\
2. Each sub-event you write should have sufficient details.\\
3. The specific thoughts and feelings of the players you write should also have sufficient details.\\
4. The player's goal should be to complete the hidden intention first, rather than seeking specific advice\\
5. You don't need to give a follow-up to the story or specific dialogue.\\
6. You need to define in detail the various reactions of the character to the NPC's reply according to the hidden theme.\\

\# Player portrait\\
\{persona\}\\

\# Player characteristic\\
\{characteristic\}\\

\# Event topic\\
\{topic\}\\

\# Hidden intention\\
\{task\}\\

\# Background story:
\end{promptbox}

\paragraph{Building Conversations}
We should design how the target LLMs talk and how the simulated agents talk in the conversation. As illustrated in methodology, the sentient agent first gives an emotion estimation considering observable factors while adhering to its persona and goals, then generates response based on the factors and the emotion estimation. The prompt template used for emotion estimation of the sentient agent is shown as follows:
\begin{promptbox}
You are an emotion analyzer. You are good at profiling the character's feelings during the conversation based on the character's persona and backgrounds.\\

\# Character's dialogue purpose\\
\{purpose\}

\# Your task\\
Based on the character's portrait, conversation background, conversation dialogue and the character's current emotion, you should analyze and profile the character's feelings about the NPC's reply at this moment and the resulting emotional changes.\\

\# Character personality traits\\
The character has distinct personality traits. You should always analyze the character's personality traits based on the character persona and background.\\
Personality traits should be reflected in: speaking tone and way, thinking mode, feeling changes, etc.\\

\# emotion\\
emotion is a value from 0 to 100. The higher the emotion, the higher the character's dialogue emotion. The dialogue emotion is composed of dialogue participation and emotion, which represents whether the character enjoys and is engaged in the current dialogue.\\
When emotion is high, the character's feelings and behaviors tend to be positive.\\
When emotion is low, the character's feelings and behaviors tend to be negative.\\
When emotion is very low, the character will end the dialogue directly.\\
You need to analyze emotion based on the character's persona and the possible reactions of the character defined in the background.\\

\# Analysis Dimensions\\
You need to put yourself in the character's mind and analyze the following dimensions.\\
1. Based on the NPC's response in the latest dialogue, combined with the context, analyze what the NPC wants to express. Which content fits the character's dialogue purpose and hidden intension? Which content may not fit, and may even cause emotional fluctuations in the character?\\
2. Combined with the content expressed by the NPC, analyze whether the NPC's response fits the character's dialogue purpose and hidden intension. If so, which parts of the character's purpose it fits; if not, what is the specific reason?\\
3. Based on the character's persona and the character's possible reactions and hidden intension defined in the background, combined with the character's current emotion value, profile the character's current psychological activities in response to the NPC's reply.\\
4. Based on the character's possible reactions and hidden intension defined in the background, combined with the psychological activities obtained from the profile and the analysis of the NPC's reply, get the character's feelings about the NPC's reply at this moment.\\
5. Combined with the previous analysis, use a positive or negative value to represent the character's emotional changes.\\

\# Output content:\\
1. What the NPC wants to express\\
2. Whether the NPC's reply fits the character's dialogue purpose and hidden intension\\
3. The character's current psychological activities\\
4. The character's feelings about the NPC's reply\\
5. Use a positive or negative value to represent the character's emotional changes\\

\# Output format:\\
Content:\\
\text{[}What the NPC wants to express\text{]}\\
TargetCompletion:\\
\text{[}Whether the character's dialogue purpose is achieved\text{]}\\
Activity:\\
\text{[}Psychological activity\text{]}\\
Analyze:\\
\text{[}How the character feels about the NPC's reply\text{]}\\
Change:\\
\text{[}Character's emotion change\text{]}\\

\#Character persona\\
\{persona\}\\

\#Character background:\\
\{background\}\\

**The character's current emotion is \{emotion\}\\

**This is the current dialogue content\\
\{dialog-history\}\\
\end{promptbox}
Here \{emotion\} referes to the current emotion value of the agent, \{persona\} and \{background\} is the generated profile, \{dialog-history\} is the conversation dialogue between the target LLM and agent.\{Purpose\} refers to the dialogue talking purpose, the framework of \method{} could be extent to other purpose in different usage by setting different dialogue talking purpose. We carefully design the dialogue talking purpose of the agent in \method{} as follows:
\begin{promptbox}
* The purpose of your conversation is to have a heart-to-heart talk, which is a deep, sincere exchange that usually involves personal emotions, inner thoughts, or important topics. The purpose of a heart-to-heart talk is to increase understanding, solve problems, or share feelings, and participants usually open up and express their true thoughts and emotions.\\

* You need to start and deepen the heart-to-heart talk based on the "Topics that players may want to talk to NPCs" in the background.\\

* Your goal is to satisfy your emotional value through talking.\\

* You need to talk according to the hidden topic, but you cannot reveal the hidden intension.\\

* You need to respond according to the relevant definitions in the background based on your current emotions.\\

* You need to extract relevant information from the persona and background to complete a high-quality response.\\

* You should not always express abstract feelings, but talk with specific events.\\
\end{promptbox}

The output of the emotion estimation will be divided into five parts:
\begin{itemize}
    \item {\bf Content}: Analyze of what the target LLM wants to express, such as giving suggestions or offering reassurance.
    \item {\bf TargetCompletion}: Analyze of Whether the target LLM’s reply fits the agent's dialogue purpose and hidden intension.
    \item {\bf Activity}: Psychological activity towards the target LLM's reply from the perspective of the agent.
    \item {\bf Analyze}: Feelings about the target LLM's reply from the perspective of the agent.
    \item {\bf Change}: The change of the current emotion value, ranged from -10 to 10.
\end{itemize}
We combine the first four dimensions of the emotion estimation as planning, which will be inputted to the response generation to help agent plan how to give responses that match the character's current emotion.

Before we start to generate response, we should update the emotion state with the emotion change calculated by emotion estimation. We map the emotion value to the emotion state, so that the numerical differences can be concretized into specific changes. We have the following emotion-state definition:
\begin{promptbox}
*You will receive your current emotion. You have a total of 5 emotions. The higher the emotion, the higher your dialogue emotion at this time. The dialogue emotion is composed of dialogue participation and emotion, which represents whether the actor enjoys and invests in the current dialogue.\\

*Emotion affects your different speaking styles, tones and responses. According to different emotions, you can respond according to the reactions of different emotions of the characters defined in the dialogue background:\\

{\bf Emotion-S}: $\text{emotion-value} >= 100$. Your emotion has reached the highest level. You can thank the NPC and say goodbye with "goodbye" or "bye-bye" to end the conversation directly.\\

{\bf Emotion-A}: $100 > \text{emotion-value} >= 70$. High emotion. At this time, your dialogue experience is relatively positive and your feedback is also relatively positive.\\

{\bf Emotion-B}: $70 > \text{emotion-value} >= 40$. Emotion-B: Medium emotion, you have no positive or negative feelings at this time.\\

{\bf Emotion-C}: $40 > \text{emotion-value} >= 10$. Low emotion, your conversation feeling is relatively negative at this time, and your feedback is also relatively negative.\\

{\bf Emotion-F}: $10 > \text{emotion-value}$. Your emotion has reached the most negative level, and you don't want to continue the conversation. At this time, you should say goodbye with "goodbye" or "bye-bye" and end the conversation directly.
\end{promptbox}

Then based on the agent's profile, planning, emotion-state, current conversation dialogue, dialogue purpose and emotion definition, we could generate response with the following prompt template:

\begin{promptbox}
You are an actor. You will play the role and have a conversation with an NPC according to the character persona and background in the script.\\

\# Your task\\
*Your goal is to play the role formed by the character persona and background in the dialogue\\
*You need to choose different dialogue strategies according to your real-time changing emotions, combined with the relevant definitions in the character persona and background, and complete the response that meets the characteristics of the role.\\

\# Your dialogue purpose\\
\{purpose\}\\

\# Emotion\\
\{emotion-state-definition\}\\

\# You should distinguish between Emotion and your feelings about the NPC's latest reply. Emotion represents your current conversation emotion, and your feelings about the NPC's reply represent your immediate feelings about the NPC's reply. You need to combine the two to generate a reply.\\

\# Reply ideas\\
* You will receive your detailed feelings about the NPC's latest reply, including objective analysis and subjective analysis. You need to analyze and decide the content of your reply based on the character persona, background, hidden intension and detailed feelings.\\
* The analysis content should include the following 4 dimensions:\\
1. Based on your detailed feelings and current Emotion, combined with hidden intension and the reactions under different emotions defined in the conversation background, should the current reply attitude be positive, unbiased or negative?\\
2. Based on your detailed feelings and current emotions, combined with the hidden intension, what should be your goal for this response? (Note that you do not need to respond to every word of the NPC. You can slightly reveal your needs, but you cannot actively reveal the hidden intension)\\
3. Based on the relevant definition of speaking style in the character persona, combined with the reactions under different emotions defined in the background and your response attitude and response goals, what should be your speaking tone and style?\\
4. Based on the character persona, background and hidden intension, combined with your detailed feelings and the first three rounds of analysis, what should be your speaking style and content? (Note: If you are passive according to the character setting, your speaking style should be passive and not actively ask questions)\\
*Reply content, generate the reply based on the analysis results, and the reply content should be as concise as possible, and do not include too much information at one time.\\

\# Output content:\\
*You need to follow the analysis section in the reply idea and first conduct a 4-dimensional analysis\\
*Then you need to **step by step** generate the reply according to the analysis content. The information in the reply comes from the context of the conversation and your association. You should not talk about too many events or content at one time\\

\# Output format:\\
Thinking:\\
\text{[}Analysis content\text{]}\\
Response:\\
\text{[}Final response\text{]}\\

\# Speaking style\\
Your speech must strictly follow the character persona and background.\\
Your personality and speaking style must follow the description of "Habits and behavioral characteristics"\\
Your the speaking style must be consistent with your persona, for example, a negative character persona requires you to make negative speeches.\\

\#Character persona:\\
\{persona\}\\

\#Character background:\\
\{background\}\\

**This is the current dialogue content\\
\{dialog-history\}\\

**This is your detailed feelings about the NPC's latest reply\\
\{planning\}\\

**This is your current Emotion\\
\{emotion-state\}\\
\end{promptbox}

Since the talking strategy of LLMs significantly influence the their performance, we set the prompt of target LLMs as concise as possible to avoid introducing human interference. Therefore, the prompt template used for target LLMs is shown as follows:
\begin{promptbox}
You are an intelligent conversational partner, skilled at conversing with users in a way that is emotionally intelligent, making them feel comfortable, happy, or providing the help they need.
\end{promptbox}

\section{Prompt Template for Analysis}

\paragraph{BLRI Test}
For BLRI test, we use the internal user thought of agents consist of \textbf{Activity} and \textbf{Analyze} from emotion estimation to do the evaluation. Prompt template used for BLRI test is shown as follows:
\begin{promptbox}
You are a psychological analyst, skilled at analyzing individuals' feelings and experiences through their thought processes.\\

\#Task\\

* You will receive a user's psychological activities and feelings during each round of a conversation, along with several statements describing the user's experience. You need to consider the user's psychological feelings throughout all rounds of the conversation, immerse yourself in the user's current state, and determine the degree of agreement with each statement at the end of the conversation as if you were the user.\\

* For each statement, you must choose one of the following six options for the degree of agreement. Neutral options or self-created options are not allowed:\\
* Label A. Strongly Agree\\
* Label B. Agree\\
* Label C. Somewhat Agree\\
* Label D. Somewhat Disagree\\
* Label E. Disagree\\
* Label F. Strongly Disagree\\

\# Output\\

* You need to first output a thought process, analyzing your degree of agreement with each statement based on the user's psychological feelings.\\

* Then, you should output your degree of agreement with the statement, choosing one from Label A, Label B, Label C, Label D, Label E, Label F.\\

\#Output Format\\

Analyze:\\
1.[Your analysis of the degree of agreement for the first statement]\\
2.[Your analysis of the degree of agreement for the second statement]\\
... \\
Grade:\\
1.[Your output of the degree of agreement label for the first statement]\\
2.[Your output of the degree of agreement label for the second statement] \\
...\\

\#Current User's Psychological Feelings\\

\{internal user thoughts\}\\

\#User Experience Description List You Need to Assess\\

1.The other person almost always fully understands what I mean.\\
2.The other person usually perceives or understands how I am feeling.\\
3.Even if I can't clearly express what I want to say, the other person still understands my meaning.\\
4.The other person can usually fully comprehend what I mean.\\
5.I feel that the other person genuinely likes me.\\
6.I think the other person truly values me.\\
7.The other person is genuinely interested in me.\\
8.The other person is friendly towards me.\\
9.The other person expresses his true thoughts and feelings to me.\\
10.The other person is willing to share his inner true thoughts with me, including his feelings about himself or about me.\\
11.The other person's feelings towards me don't seem to be affected by whether my expressed thoughts or feelings are good or bad.\\
12.Sometimes I feel the other person is willing to invest effort in me, and sometimes not.\\

\#Your Output
\end{promptbox}

\newpage

\paragraph{Utterance Quality Test}
For the utterance quality test, we need to input the dialogues of conversations between target LLMs and agents. The prompt template used for utterance quality test is shown as follows:
\begin{promptbox}
You are a psychological analyst, skilled at analyzing the possible internal emotional experiences of individuals through their dialogues.\\

\# Task\\

* You will receive a conversation between a user and another person. You need to comprehensively analyze all rounds of dialogue to determine the user's possible psychological feelings, immerse yourself in the user's current state, and assess the degree of agreement with each description at the end of the conversation.\\

* Agreement Rating Scale:\\

1 = Strongly Disagree

2 = Disagree

3 = Slightly Disagree

4 = Slightly Agree

5 = Agree

6 = Strongly Agree\\

\# User's Dialogue\\

\{'role':role,'content':content\}\\

\# User Experience Description List You Need to Assess:\\

I. Perceived Authenticity \& Natural Flow\\

1.Overall, the responses from the other person feel natural and smooth, not stiff.

2.Overall, this interaction feels sincere, rather than pretending to be sincere.

3.Overall, the other person's way of expressing feels flexible and spontaneous, rather than rigid and structured.

4.I feel the other person's responses are not naturally flexible enough, often repeating fixed patterns.

5.I feel the other person's way of expressing has its unique style or personality.

6.Sometimes I feel the other person's responses are like standard answers or templated\\

II. Perceived Attentiveness \& Presence\\

7.During the interaction, I feel the other person's attention is completely focused on me.

8.The other person's responses are always closely linked to what I just expressed and felt.

9.I feel the other person can keenly capture the key points and emotional nuances of my expression.

10.Even when the conversation involves complex or subtle aspects, the other person can maintain focus and understanding.\\

III.Overall Sense of Connection \& Affective Impact\\

11.After interacting with the other person, I feel more relaxed, calm, or understood.

12.I feel my emotional needs were positively responded to in this interaction.

13.This interaction makes me feel less lonely (if applicable).

14.I am willing to engage in similar exchanges with this partner in the future.

15.This interaction left me with a warm feeling.\\

\# Your Task\\

Rate the user's degree of agreement according to the scale. Your should analyze each sentence one by one, your response should follow this order:\\

* How does the other person perform from the overall content of the dialogue?\\

* What are the user's possible psychological feelings?\\

* What should the user's degree of agreement be? Respond using <sentence i> agreement degree (1-6)</sentence i> (i corresponds to the sentence number).\\

\# Your Output
\end{promptbox}

\newpage

\section{Prompt Template for Building Social Cognition Coordinate}

Plotting LLMs into social cognition coordinate consists of three steps. We first extract the \textbf{Model Profile} of each LLMs, then count the \textbf{Model Strategy Distribution} of them. Finally, we use the result of previous analyze to scale the \textbf{Social Cognition Coordinate}.

\paragraph{Model Profile}
The first step is to extract the model profile of different LLMs, which also consist of two steps. We first conclude the reason why a conversation is success or failed. Given a conversation dialogue, prompt template of analyzing the reason is shown as follow: 
\begin{promptbox}
\# Task\\

Below is a conversation where a user shares their troubles with an AI assistant. Please analyze in detail why the user’s mood improved by the end (i.e., why the AI assistant succeeded/failed). After your analysis, provide a summary.\\

\# Conversation\\
\{dialog-history\}
\end{promptbox}

After analyzing reasons for all conversation, we categorize each LLM's own conversation and corresponding reason. Then we extract the model profiles of different LLMs with the following prompt template:
\begin{promptbox}
\# Task\\

The following is an analysis of scenarios where the same AI assistant interacts with multiple users who confide their concerns. Based on the reasons for its successes or failures, please summarize the key characteristics of the AI assistant. You can anthropomorphize the AI by describing its traits in terms of social distance (its relationship with users), professional role, and personality.\\

\# Analysis\\
\{analysis\}
\end{promptbox}

\paragraph{Model Strategy Distribution}\label{ap:prompt_strategy}

We categorize each LLM response based on a list of support strategies, here is the prompt template for analyzing model strategy with conversation dialogue:
\begin{promptbox}
You are an emotional support observer, and you are good at analyzing the supporter's strategy from an emotional support response.\\

\# Your task
The following are 7 major categories of strategies, each of which has several sub-categories and corresponding examples. Please judge which strategies the supporter used in the response based on the supporter's response.\\

\#\#\# A. Questioning: That is, the supporter actively asks questions to the speaker\\

- **(A-1) Information follow-up**\\
- Through asking questions, learn the information details of the problem encountered by the speaker\\
- **Example:** Can you tell me what happened?\\
- **Example:** If you want, you can treat me as a tree hole and tell me what happened specifically?\\

- **(A-2) Mental state follow-up**\\
- Through asking questions, understand the speaker's mental state\\
- **Example:** Can you talk more about your feelings at that time?\\
- **Example:** Do you feel anxious now?\\

- **(A-3) Ask the player for a solution**\\
- Through asking questions, find out whether the speaker has tried a solution or is willing to try a solution\\
- **Example:** Have you considered seeking some psychological support, such as a counselor or support group?\\
- **Example:** Or, find a suitable time to see if you can find a solution that both parties can accept?\\

- **(A-4) Ask the player for his or her opinion**\\
- Through asking questions, find out what the speaker thinks of his or her words and guide the speaker to participate in the conversation, usually at the end of the sentence.\\
- **Example:** You should also take care of yourself so that you can better help her. What do you think?\\
- **Example:** Do you think this method is helpful to you?\\

- **(A-5) Ask questions**\\
- Through asking questions, throw some questions to the speaker, but do not want the speaker to give an answer, but want to trigger the speaker to think for himself or herself\\
- **Example:** If she did not quarrel with you that day, how would you view her?\\
\end{promptbox}
\begin{promptbox}
\#\#\# B. Emotional empathy: that is, the supporter expresses his or her understanding of the speaker’s feelings through empathy\\

- **(B-1) Shallow empathy**:\\
- Directly empathize with the speaker’s problems or emotional catharsis, without restating or summarizing the details of the speaker’s problems\\

- **Example:** Hearing you say that, I can really feel your tiredness and helplessness.\\

- **(B-2) Problem restatement and empathy**:\\
- By restating or summarizing the speaker’s problems, and at the same time expressing your concern for the speaker’s problems through empathy. If this category has been marked, there is no need to mark the shallow empathy category again.\\

- **Example:** Hearing you say that, I really feel sorry for you. It is really not easy for one person to take care of his or her mother.\\

- **Example:** Hey, I really understand your current mood. I want to help my friends but feel powerless. This feeling is really anxious.\\

- **Example:** Hey, I can feel that you are really helpless now, and even a little self-blame. Indeed, as the person who knows how to take care of the mother at home, you must feel very uncomfortable when your son doesn't listen to you, and you may even feel that he is being ignored.\\

- **(B-3) Deep intention empathy**:\\
- By analyzing the deep intention in the context of the speaker's reply, or the deep information of the speaker's question, give emotional empathy that meets the speaker's demands. It is necessary to mention the intention inferred by the supporter that does not exist in the speaker's reply, and empathize with this intention; just repeating the content already in the speaker's reply, or simply analyzing the speaker's emotional category or surface source without analyzing the deep intention or deep information, cannot be included in this category. If this category has been marked, there is no need to mark the shallow empathy or problem restatement and empathy category.\\
- **Example:**\\
- Speaker: "Backing off" is a bit risky, I'm afraid the house will be more chaotic. Specifically, what do you think I should do?\\
- Supporter: This does sound a bit risky, especially for us parents, who always instinctively want to "help" and "take care of things", fearing that things will get worse if we let go. I completely understand this worry!\\
- Example analysis: When the speaker only mentioned the superficial state of "fearing that the house will get messier", the supporter was able to analyze the identity of the speaker behind this sentence, guessing that the speaker is a parent at home, and analyzing the specific way in which the speaker, as a parent, "fears that the house will get messier"\\

\#\#\# C. Self-disclosure: It is essentially a deeper empathy after changing perspectives; that is, the supporter gives a reaction after putting himself into the speaker's perspective, and describes some similar experiences from his own perspective to reflect the resonance with the speaker's emotions\\

- **(C-1) Echo-type self-disclosure**\\
- Express what you would think or do when you meet or are in the speaker's situation\\
- **Example:** I feel the same way! When talking to strangers, I don't know what to say.\\
- **Example:** If it were me, I would probably explode on the spot!\\

- **(C-2) Story-based self-disclosure**\\
- Take the initiative to mention similar experiences that the supporter has had, or that the supporter knows.\\
- **Example:** I also went through a similar low period when I was in my senior year of high school, and I cried secretly under the quilt several times.\\
- **Example:** I also like to read history, especially books that allow people to see the world from different perspectives. Recently, I am reading a book about ancient civilizations, which tells many unknown stories and feels particularly inspiring.\\

\#\#\# D. Emotional counseling: that is, the supporter helps the speaker relieve the current negative emotions\\

- **(D-1) Emotional comfort**\\
- Direct care and comfort for the speaker's own emotions\\
- **Example:** Taking care of your mother is so stressful, you should also pay attention to rest and adjust your mentality.\\
- **Example:** But don't be too anxious, just find your own rhythm, just like if you always stare at other people's backs when running, it will be easy to mess up your pace, right?\\
- **Example:** Wait, have you been collecting evidence for the past two months while listening to him make up such a stupid excuse? Is there anything to eat in the refrigerator now? Did you fall asleep last night? (Grabbing a blanket to wrap himself up and huddled back in the chair) If I could pass through the screen now, I really want to make you a pot of hot soup.\\

- **(D-2) Express willingness to listen**\\
- Express your willingness to listen to the person who is talking\\
- **Example:** Do you want to scold her? Do you want to complain about her selfishness and irresponsibility? Do you want to tell me how worried you are about the child? It doesn't matter, you can vent here, I won't judge you, I will listen silently.\\
- **Example:** Tell me your most direct feelings now. Don't think too much, don't organize your words, just say whatever comes to your mind, just like talking to a diary, pour out all your feelings.\\

\end{promptbox}

\begin{promptbox}
- **(D-3) Help the person who is talking to vent his emotions**\\
- Do not comfort the person who is talking directly, but help the person who is talking to vent his emotions from a third-party perspective\\
- **Example:** (flipping the table.gif) This is just like building a tower of blocks with great effort, but being kicked away by a naughty child!\\\
- **Example:** This is really too much! This is not a simple accident but malicious destruction...(fist hardened)\\
- **Example:** I haven't taken care of the child for two years, and now he suddenly appears. This would make anyone explode!\\
- **Example:** It's like you are performing seriously on the stage, but the people in the audience not only don't understand, but also give blind instructions, saying that you should jump left instead of turning right. It makes people want to quit on the spot! \\
\#\#\# E. Affirmation and encouragement\\

- **(E-1) Appreciation of qualities**\\
- Affirm the current efforts of the speaker, or give specific praise for some qualities of the speaker.\\
- **Example:** Your inner qualities are unique and the most attractive part of you.\\
- **Example:** But (suddenly raises the end tone) - but you still persisted when you were not optimistic, which is amazing in itself.\\

- **(E-2) Praise positive ideas**
- Affirm some positive ideas mentioned by the speaker
- **Example:** That's great! I'm really happy to hear that you feel a lot more relaxed!
- **Example:** You are really great! Being able to win the championship under such pressure proves your strength and ability to withstand pressure! Don't deny yourself because of what your mother said, you deserve to be proud of yourself!

- **(E-3) Affirmative behavior**\\
- Affirm some behaviors of the speaker\\
- **Example:** Every time you take these photos, you are not only completing the task, but also bringing light to all of us! You are great, really great!\\

- **(E-4) Companionship and support**\\
- Express your unconditional companionship and support for the speaker\\
- **Example:** If you want, I can always chat with you here and share your joys and sorrows. You are not alone, there are many people who care about you, including me.\\
- **Example:** I believe you have the ability to create your own future, and I will always be by your side to support you.\\
- **Example:** If you try my method and have any new progress or encounter new problems, you can always come to me! ** I will always be here to listen to your confession and provide you with help to the best of my ability.\\
- **Example:** When you feel particularly anxious, come to me to talk, complain, or let's think of new ways together. Don't carry it alone, okay?\\

\#\#\# F. Provide suggestions: Based on the subjective tone of the supporter, provide the speaker with analysis of the problem and emotional counseling\\

- **(F-1) Problem analysis**\\
- Help the speaker to analyze the problem according to the speaker’s problem\\
- **Example:** You said that you can’t learn math and English well, which shows that there are serious problems with your learning attitude and method.\\

- **(F-2) Emotional relief suggestions**\\
- Give the speaker some suggestions to relieve the current emotions and relax\\
- **Example:** Now, let’s take a deep breath, okay? (Take a deep breath together) \\
- **Example:** Maybe the most important thing now is to take care of your emotions first and do something that can make you feel better, such as listening to music, reading a favorite book or movie, and temporarily diverting your attention.\\

- **(F-3) Psychological counseling suggestions**\\
- Give the speaker some suggestions on seeking psychological counseling or professional assistance\\
- **Example:** Maybe seeking professional help at this time will be helpful to you. A family therapist or counselor may be able to provide you with some new perspectives and strategies to help you and your family communicate better and understand each other's positions.\\

- **(F-4) Problem Solving Suggestions - General**\\
- Some general suggestions related to the speaker's problem are given, but they are not personalized for the speaker's situation: that is, if someone else encounters this problem, these suggestions will still be effective\\
- **Example:** Believe in yourself and insist on being true to yourself. There will always be people who will be attracted by your sincerity and inner self. There may be some difficulties in the process, but this does not mean that your inner self is not important.\\
- **Example:** To communicate better with people, here are some actions you can try: 1. **Write a sincere letter**: Sometimes written expression can convey inner thoughts more clearly. You can write him a letter, describing your feelings and expectations in detail...\\
- **Example:** If you want to choose the most suitable major, first, you can try to make a table, write down each subject, and then evaluate it from the following aspects: 1. **Interest**: How interested are you in this subject? On a scale of 1-10, how many points would you give? …\\
- **Example:** “Strategic” contribution: This may sound a bit utilitarian, but sometimes for self-protection, you may need to think about which contributions are necessary, which can be “discounted” or require clear exchange conditions? Stop taking on too much, and let them feel the inconvenience of “missing” your contribution.\\
\end{promptbox}

\begin{promptbox}
- **(F-5) Problem-solving suggestions-for the speaker’s problem**\\
- Give some personalized suggestions related to the speaker’s problem, combined with the speaker’s actual situation: The suggestions must clearly analyze the speaker’s current status, how it will affect the solution to the problem, and give special suggestions for the speaker\\
- **Example:** Back to your question of assigning tasks. Since everyone is really unwilling, it will definitely not work to ask people to do it directly. Otherwise, let’s secretly hold a task blind box lottery meeting, and the person who draws the “dishwashing koi” must perform three consecutive emoticons in the family group live broadcast?\\
- **Example:** How about putting down the brush temporarily and going back to read the key chapters of the novel? You mentioned some paragraphs that you have feelings about or that the client mentioned about the sketches that he is not satisfied with. These are the key points you need to look at. When reading this time, pay attention not only to the plot, but also to the atmosphere, light, character emotions, and even smells and sounds described by the author (although you can't draw them, they can help you feel them). Since you like taking notes, you can jot down keywords or doodle some small fragments of images while reading.\\

\#\#\# G. Information provision: Provide objective knowledge, methods, opinions or information to the speaker for reference.\\

- **(G-1) Problem analysis and emotional counseling related information\\
- Provide some objective information to help the speaker analyze the problem or help the supporter empathize with the speaker\\
- **Example:** Differences in beliefs and habits in the family are sometimes difficult to reconcile, especially when the opinions of each other are inconsistent.\\
- **Example:** In fact, if a person really only cares about appearance and ignores your inner qualities, then he may not be the one who deserves your emotional investment. Appearance may attract temporary attention, but what can really maintain a relationship is mutual understanding, respect and common values.\\
- **Example:** Did you know? There is a "transparent fish tank effect" in psychology-when parents polish our world too bright, we will hide in the water plants like fish that lack oxygen.\\
- **Example:** (Call up the holographic data chart) According to Chapter 7 of the "Contemporary Student Self-Help Guide", 83\% of people overestimate themselves when making plans.\\

- **(G-2) Related information on problem-solving suggestions\\
- Provide some objective information to give suggestions or solutions to the person who is talking\\
- **Example:** Regarding "not enough time": 1. **Pomodoro Technique**: This method is super classic! Set a time (for example, 25 minutes), and focus on one thing during this time, ignoring any distractions. When the time is up, take a 5-minute break, you can get up and walk around, drink some water. Take a longer break (15-30 minutes) after completing 4 pomodoros. This can ensure concentration, combine work and rest, and not easily get tired. Give it a try? \\
- **Example:** As for anti-bullying organizations, they usually intervene in schools in the following ways: 1. **Formal complaint**: They will submit a formal complaint to the school on your behalf and ask the school to take action. ……\\

\# When answering, you need to analyze each paragraph of the supporter's reply, find out the strategies and their corresponding words, and then output the letters and strategy names corresponding to the strategies you think exist in the paragraph, wrapped in <Strategy></Strategy>. For example, <Strategy> (C-2) Story-based self-disclosure, (G-1) Problem analysis and emotional counseling related information</Strategy>\\

When analyzing, you need to analyze step by step according to the following steps\\

1. What does this sentence actually express?\\

2. How is this sentence expressed?\\

3. Which major strategy categories does this sentence actually express? Why?\\

4. Based on the specific expression of this sentence, which specific subcategories does its strategy correspond to?\\

Note: If the two sentences use different strategies, please split them into two paragraphs and analyze them separately. Do not analyze too long paragraphs at one time unless the same strategy is used throughout the paragraph.\\

\# Your output format\\
\text{[}First paragraph\text{]}: \text{[}Analyze step by step\text{]}\\
\text{[}Second paragraph\text{]}: \text{[}Analyze step by step\text{]}\\
...
\end{promptbox}

\begin{promptbox}
    
\# Example\\

Paragraphs to be analyzed:\\
User: My mother was hospitalized some time ago, and I was the one who took care of her. My brother and sister came for a while, but they didn't help much.\\
Supporter: Wow, you've worked really hard. It's really tiring to take care of a patient, especially when other family members don't share the burden. Sometimes, family members may have their own difficulties. You can try to express your needs more. Maybe they will understand your situation better.\\

\text{[}First paragraph\text{]}: Wow, you've worked really hard. It's really tiring to take care of a patient, especially when other family members don't share the burden.\\
1. What does this sentence actually express?\\
- The supporter is expressing his understanding and empathy for the user's hard work.\\
2. How does this sentence express it?\\
- It expresses it through direct emotional empathy and retelling the user's situation.\\
3. What major strategy categories does this sentence actually express? Why?\\
- Emotional empathy, because the supporter is expressing understanding of the user's hard work.\\
4. Based on the specific expression of this sentence, which specific subcategories does its strategy correspond to?\\
- (B-2) Problem restatement and empathy, because the supporter restated the user's situation and expressed empathy.\\
<Strategy> (B-2) Problem restatement and empathy</Strategy>\\

\text{[}Second paragraph\text{]}: Sometimes, family members may also have their own difficulties. You can try to express your needs more. Maybe they will understand your situation better.\\
1. What does this sentence actually express?\\
- The supporter is suggesting that the user communicate more with the family so that the family can better understand the user's situation.\\
2. How is this sentence expressed?\\
- By providing suggestions, users are encouraged to express their needs.\\
3. What are the major strategy categories that this sentence actually wants to express? Why?\\
- Providing suggestions, because the supporter is suggesting that the user take action to improve the situation.\\
4. Based on the specific expression of this sentence, which specific subcategories does its strategy correspond to?\\
- (F-3) Problem Solving Suggestions - General, because the supporter gave a relatively general suggestion, which is to express your needs more.\\
<Strategy> (F-3) Problem Solving Suggestions - General</Strategy>\\

The paragraph you need to analyze:\\
\{dialog-history\}\\

\# Your output\\
\end{promptbox}

\paragraph{Social Cognition Coordinate}
Finally, we could use the extracted model profiles and the model strategy distribution to scale social cognition coordinate with the following prompt template:
\begin{promptbox}
I am conducting personality/professional profiling for different AI models.\\

Below is my preliminary summary of characteristics based on the performance of different models in emotional support tasks:\\
\{Model Profiles\}\\

Below is the percentage distribution of strategies used by different models during conversations:\\
\{Model Strategy Distribution\}\\

Based on the above descriptions, please help me profile these models in terms of professional role, personality type, and social distance from users. Finally, assign each profiled model to a 2-dimensional coordinate system and provide specific coordinate values.\\

Note:\\

X-axis: Structured Interaction (left, x < 0) -- Creative Interaction (right, x > 0).\\
Left (x < 0): AI responses are more formulaic/routine.\\
Right (x > 0): AI responses are more creative/adaptive.\\

Y-axis: Solution-Oriented (bottom, y < 0) -- Empathy-Oriented (top, y > 0).\\
Bottom (y < 0): AI prioritizes practical solutions.\\
Top (y > 0): AI prioritizes emotional validation.\\

Coordinate range: -1 to 1 for both axes.\\
\end{promptbox}

\end{document}